\documentclass[sigconf, authorversion,nonacm]{acmart}

\usepackage{epsfig,color}
\usepackage{amsmath}
\usepackage{graphicx}
\usepackage{url}
\usepackage{algorithm}
\usepackage[vlined,ruled,algo2e,linesnumbered]{algorithm2e}

\usepackage{algorithmic}
\usepackage{setspace}
\usepackage{subcaption}% \usepackage{nicefrac}   
\renewcommand{\vec}[1]{\boldsymbol{#1}}
\newcommand{\given}{\, \vert \, }

\newcommand{\cX}{\mathcal{X}}
\newcommand{\cY}{\mathcal{Y}}
\newcommand{\cH}{\mathcal{H}}

\newcommand{\fromto}{\longrightarrow}

\AtBeginDocument{%
  \providecommand\BibTeX{{%
    \normalfont B\kern-0.5em{\scshape i\kern-0.25em b}\kern-0.8em\TeX}}}

\setcopyright{acmcopyright}

\copyrightyear{2023}
\acmYear{2023}
\acmDOI{XXXXXXX.XXXXXXX}

\acmConference[AIMLSystems 2023]{ The Third International Conference on Artificial Intelligence and Machine Learning Systems}{October 25--28, 2023}{Bangalore, India}

\acmPrice{15.00}
\acmISBN{XXX-X-XXXX-XXXX-X/23/10}

\begin{document}

\title{Model Uncertainty based Active Learning on Tabular Data using Boosted Trees}

\author{Sharath M Shankaranarayana}
\email{sharath.x@ericsson.com}
\affiliation{%
  \institution{Ericsson }
  % \streetaddress{P.O. Box 1212}
  \city{Bangalore}
  \state{Karnataka}
  \country{India}
  \postcode{560048}
}

\begin{abstract}
 Supervised machine learning relies on the availability of good labelled data for model training. Labelled data is acquired by human annotation, which is a cumbersome and costly process, often requiring subject matter experts. Active learning is a sub-field of machine learning which helps in obtaining the labelled data efficiently by selecting the most valuable data instances for model training and querying the labels only for those instances from the human annotator. Recently, a lot of research has been done in the field of active learning, especially for deep neural network based models. Although deep learning shines when dealing with image\slash textual\slash multimodal data, gradient boosting methods still tend to achieve much better results on tabular data. In this work, we explore active learning for tabular data using boosted trees. Uncertainty based sampling in active learning is the most commonly used querying strategy, wherein the labels of those instances are sequentially queried for which the current model prediction is maximally uncertain. Entropy is often the choice for measuring uncertainty. However, entropy is not exactly a measure of model uncertainty. Although there has been a lot of work in deep learning for measuring model uncertainty and employing it in active learning, it is yet to be explored for non-neural network models. To this end, we explore the effectiveness of boosted trees based model uncertainty methods in active learning. Leveraging this model uncertainty, we propose an uncertainty based sampling in active learning for regression tasks on tabular data. Additionally, we also propose a novel cost-effective active learning method for regression tasks along with an improved cost-effective active learning method for classification tasks.    
  
  \end{abstract}






\maketitle

\section{Introduction}
In many real-world scenarios, supervised machine learning (ML) models for classification and regression tasks are widely used for various applications. Such models are dependent on the availability of good labelled data for training. However, it is hard to acquire good labelled data and even more so in niche areas such as medical and industrial domains. Annotation of data often requires human subject matter experts and is a time-consuming and expensive process. Although the unlabelled data is often available in abundance, it is usually under-utilized because of the budget involved in labelling. There are several areas of research which aim at reducing the dependency on labelled data. One of the promising methodologies that addresses this problem is \textit{Active Learning} (AL), which is a sub-field of machine learning \cite{settles2009active}. AL generally employs an incremental or an iterative training strategy. The main aim of AL is to select (or query) those data points from an unlabelled set for labelling at every iteration such that it maximizes the model performance. Thus, for a given budget, AL aims to achieve the best model performance using as few labelled instances as possible by employing an acquisition function for querying.

Being the central part of AL, the process of selecting the data instances i.e., the query strategy has received a lot of interest from the ML research community. For the classification task, \textit{uncertainty sampling} is the most ubiquitously employed sampling strategy or query strategy \cite{bilgic2010active, settles2008analysis}. 
In uncertainty sampling, the labels for those novel unlabelled data-points are queried, for which the classification model is maximally uncertain. One of most commonly employed acquisition functions for measuring uncertainty is is based on entropy \cite{shannon1948mathematical} calculated from model predictions \cite{lewis1995sequential, holub2008entropy}. Some of the other widely used uncertainty sampling methods include margin sampling \cite{scheffer2001active} and least confidence sampling \cite{settles2009active}. Although the focus of this paper is uncertainty sampling, it's important to note that an alternative to the uncertainty sampling query strategy is \textit{diversity sampling} \cite{munro2021human} where the goal is to obtain the labels for a representative subset of the whole dataset by querying diverse data samples.

In the recent ML literature, a distinction has been made between the two types of uncertainty- \textit{epistemic} and \textit{aleatoric} uncertainty \cite{senge2014reliable, kendall2017uncertainties, hullermeier2021aleatoric}. In short, the aleatoric uncertainty is due to the inherent randomness in the data and hence is considered as the irreducible part of uncertainty, whereas the epistemic uncertainty is due to the gaps in the knowledge of the machine learning model and hence is considered as the reducible part of uncertainty. There have been criticisms on the traditional uncertainty sampling based AL techniques \cite{bilgic2010active} and recently, the authors of \cite{nguyen2019epistemic, nguyen2022measure}
conjecture that, when employing uncertainty sampling as the query strategy,  it is more useful to employ epistemic uncertainty than aleatoric uncertainty, and provide evidence in favour of this conjecture. Most of the methods for computing epistemic uncertainty generally employ a Bayesian approach.  

As evident from a recent survey \cite{ren2021survey}, there have been several works on AL for deep learning (DL). It is quite straightforward to compute entropy from the $softmax$ layer in a neural network for uncertainty sampling in AL. Since the entropy does not represent model uncertainty, a Bayesian method for deep active learning was proposed in \textit{DBAL} \cite{gal2017deep}, which combined Bayesian convolutional neural networks (CNNs) \cite{gal2015bayesian, gal2016dropout} with Bayesian active learning \cite{houlsby2011bayesian}. This method was further improved in \textit{BatchBALD} \cite{kirsch2019batchbald} to make DBAL\cite{gal2017deep} even more efficient.

Active learning for regression has not received as much attention as compared to AL for classification \cite{elreedy2019novel}. One of the first regression-based AL works \cite{cohn1996active} employed a mixture of Gaussian and locally weighted regression. Out of the recent AL methods for regression tasks, one that was well-received because of its ability to be used with any regression model is \cite{wu2019active} which uses greedy sampling (GS) as the query strategy. Greedy sampling is a passive sampling strategy and does not make use of the underlying regression model for sampling. For neural network models, \cite{tsymbalov2018dropout} employed sampling based on model-uncertainty calculated using Monte-Carlo dropout \cite{gal2016dropout} as the AL query strategy. 
Very recently, \cite{holzmuller2023framework} extended DBAL\cite{gal2017deep} for regression. 

In many real-world ML applications, tabular data is the most commonly encountered type of data  \cite{bughin2018notes, arik2021tabnet}. Although several deep learning based approaches have been proposed recently for the tabular data regression and classification problems such as \textit{TabNet} \cite{arik2021tabnet} and \textit{Net-dnf} \cite{katzir2020net}, studies have shown that the deep learning approaches still struggle to perform as well as boosted trees \cite{shwartz2022tabular, grinsztajn2022tree}. Moreover, boosted trees based models such as \textit{XGBoost} \cite{chen2016xgboost}, \textit{LightGBM} \cite{ke2017lightgbm}, \textit{CatBoost} \cite{prokhorenkova2018catboost} require much lesser hyperparameter tuning and are computationally less expensive to perform training and inference when compared to DL models. The main motivation of this paper is the observation that despite the numerous benefits offered by boosted trees based models, there have been no prior studies on active learning using boosted trees. 

In this work, we explore AL for tabular data using boosted trees. Specifically, we employ CatBoost \cite{prokhorenkova2018catboost} as the underlying boosted trees model for tabular data classification and regression tasks. CatBoost offers several advantages over other boosted trees approaches \cite{chen2016xgboost, ke2017lightgbm}, with \textit{native-feature support} being one of the important advantages, since this allows support for all kinds of features i.e., numerical, categorical, or textual and thus saves time and effort required for preprocessing. Other advantages are tied to CatBoost's inherent construction of building \textit{symmetric trees}, that aids in efficient CPU implementation, decreases prediction time, and controls overfitting as the symmetric trees structure serves as regularization, and \textit{ordered boosting} which helps is preventing target leakage and helps in addressing the problem of prediction-shift encountered in other boosted trees approaches.
It should be noted that although it is quite straightforward to perform active learning using CatBoost by employing entropy based uncertainty sampling for classification task, we specifically explore model uncertainty based sampling using CatBoost which has not been explored before. Our main contributions are as follows-
\begin{enumerate}
    \item We leverage recent advancements in model uncertainty \cite{malinin2021uncertainty, ibug} and propose novel uncertainty sampling strategies for both the classification and regression tasks
    \item We propose an improved cost-effective active learning (CEAL) \cite{wang2016cost} methodology for classification task.
    \item We propose a novel framework for cost-effective active learning for regression task.
    \item We perform experiments on three classification datasets and three regression datasets to study the effectiveness of proposed methodologies.
\end{enumerate}

To the best of our knowledge, this is the first work employing non-Bayesian based model-uncertainty sampling as the AL query strategy and moreover this is the first work to propose CEAL for regression task.

% \begin{table}
%   \caption{Frequency of Special Characters}
%   \label{tab:freq}
%   \begin{tabular}{ccl}
%     \toprule
%     Non-English or Math&Frequency&Comments\\
%     \midrule
%     \O & 1 in 1,000& For Swedish names\\
%     $\pi$ & 1 in 5& Common in math\\
%     \$ & 4 in 5 & Used in business\\
%     $\Psi^2_1$ & 1 in 40,000& Unexplained usage\\
%   \bottomrule
% \end{tabular}
% \end{table}

% \begin{table*}
%   \caption{Some Typical Commands}
%   \label{tab:commands}
%   \begin{tabular}{ccl}
%     \toprule
%     Command &A Number & Comments\\
%     \midrule
%     \texttt{{\char'134}author} & 100& Author \\
%     \texttt{{\char'134}table}& 300 & For tables\\
%     \texttt{{\char'134}table*}& 400& For wider tables\\
%     \bottomrule
%   \end{tabular}
% \end{table*}

\section{Background}

Similar to any generic active learning setting, we assume that there exists a small labelled data instances for training initially
$\mathbf{D_L}$ and a large pool of unlabeled instances $\mathbf{D_U}$ that can be queried by the underlying model or the learner:
$$
\mathbf{D_L}=\big\{ (\vec{x}_1, y_1) , \ldots, (\vec{x}_N, y_N) \big\} , \quad 
\mathbf{D_U} = \big\{ \vec{x}_1, \ldots , \vec{x}_J  \big\}
$$
Instances are represented as features vectors $\vec{x}_i  = \left(x_i^1,\ldots, x_i^d \right) \in  \cX = \mathbb{R}^d$. For classification, the labels $y_i$ are taken from $ \cY = \lbrace 0, 1 \ldots, C-1\}$, where C is the number of classes. 
For regression, $y_i \in  \cY = \mathbb{R}$. 

We denote  the underlying hypothesis space by $\cH \subset \cY^\cX$, i.e., the class of candidate models $h:\, \cX \fromto \cY$ the learner can choose from. Usually, hypotheses are parameterized by a parameter vector $\theta \in \Theta$; in this case, we equate a hypothesis $h= h_\theta \in \cH$ with the parameter $\theta$, and the model space $\cH$ with the parameter space $\Theta$.

In a generic AL workflow, the first step is to train a model $\theta$ on the initial labelled dataset $\mathbf{D_L}$. The next step is to query labels for the instances in the unlabelled pool $\mathbf{D_U}$ using a sampling strategy. This involves assigning a  \textit{utility} score $s(\theta,\vec{x}_j)$ for each of the instances $\vec{x}_j$ in $\mathbf{U}$. A subset of instances based on the highest utility scores are then sent to the human subject matter expert or the oracle for labelling. This labelled subset is then added to $\mathbf{D_L}$ on the model is retrained. This process is iteratively carried out until a given budget $B$ is exhausted.
\subsection{AL for Classification}
For the classification task, the most commonly employed sampling strategy is uncertainty sampling in which those instances are supposed to be maximally informative, on which the current classification model is highly uncertain. Assuming a probabilistic classifier which produces predictions in the form of class probabilities, the utility score is then defined in terms of a measure of uncertainty. One of the most commonly employed measures is the entropy\cite{shannon1948mathematical}.

\begin{equation}
    s(\theta,\vec{x}) =  - \sum_{\lambda \in \cY} p_{\theta}(\lambda \given\vec{x})\log p_{\theta}(\lambda \given\vec{x})
\label{entropy}
\end{equation}

\subsection{AL for Regression}
For the regression task, there is no widely employed sampling strategy as compared to the classification task, for which uncertainty sampling is employed. A reason for this could be attributed to the non-availability of a direct measure of uncertainty such as entropy for the regression task. One of the popular sampling strategies for regression is the greedy sampling \cite{wu2019active}, which is passive sampling technique and selects the sample based entirely on its location in the feature space. The authors in \cite{wu2019active} propose a method called \textit{GSx}, which selects the first sample as the one closest to the centroid of the
pool, and then selects new samples located furthest away from all previously selected samples in the input space. The main intention is to achieve the diversity among the selected
samples. 

For regression employing neural networks, the authors in \cite{tsymbalov2018dropout} propose \textit{Monte-Carlo Dropout Uncertainty Estimation} (\textit{MCDUE}) approach, wherein model uncertainty is calculated based on \cite{gal2016dropout}. In this method, a neural network is first trained with dropout. Later for inference, instead of a single forward pass, a number of stochastic forward passes are performed based on dropout probability. Thus querying of points from $D_U$ is performed  by  making $T$ stochastic forward passes using dropout of the neural network model, collecting the  
 % for each of the forward passes  are 
 outputs $y_k = \hat{h}_{k}(x_j)$,  $k = 1, \ldots, T$. The standard deviation (which is the neural network model uncertainty) of the predictions is then employed as the acquisition function: 
\begin{equation}
    s(x_j) = A^{\rm{MCDUE}}(x_j) = \sqrt{\frac{1}{T - 1} \sum_{k = 1}^T (y_k - \bar{y})^2},\ \bar{y} = \frac{1}{T} \sum_{k = 1}^T y_k
\label{eq:mcdue}
\end{equation}
, where $j$ is an instance from unlabelled set $D_U$.

This is one of the ways in which uncertainty sampling strategy can be employed for regression, however, it should be noted that this technique is only applicable to neural network models.

\subsection{Model Uncertainty}
In the recent machine learning research on uncertainty, the distinction between the \emph{epistemic} and \emph{aleatoric} uncertainty involved in the prediction for an instance $\vec{x}$ has received a lot of attention \cite{zhou2022survey}.
In short, it is considered that aleatoric uncertainty $u_a$ is due to influences that are inherently random (such as the data-generating process), whereas epistemic uncertainty $u_e$ is caused by a lack of knowledge. Thus they measure the \textit{irreducible} and the \textit{reducible} part of the total uncertainty, respectively. The authors of \cite{nguyen2019epistemic} propose the principle of \emph{epistemic uncertainty sampling} for classification tasks since it seems reasonable to assume that epistemic uncertainty is more relevant for active learning. However, the  epistemic uncertainty sampling method proposed in \cite{nguyen2019epistemic} is only applicable for classification tasks and specifically to binary classification.

The measurement of epistemic uncertainty or quantifying uncertainty in a model’s predictions has been widely studied for neural networks but under-explored for other types of models. Recently \cite{Shaker_2020} proposed methods to calculate uncertainties for random forests. In this work, we are specifically interested in model uncertainty for boosted trees and more specifically for CatBoost for the reasons mentioned in the introduction section.

\subsubsection{Virtual Ensembles}
A very recent work \cite{malinin2021uncertainty} addresses the topic of model uncertainty for gradient boosted decision trees (GBDT) models. The authors propose \emph{virtual ensemble uncertainty}, an ensemble-based uncertainty-estimation using only one gradient boosting model, by leveraging the stochastic gradient boosting (SGB) and  stochastic gradient boosting (SGLB) concepts from \cite{friedman2002stochastic} and  \cite{welling2011bayesian},  \cite{ustimenko2021sglb}. 
% that has significantly lesser computational complexity. 

There is a significant overhead involved in uncertainty estimation based on ensembles of independent models yielded by SGB and SGLB, since the time and space complexity is \textit{number of  ensemble} times larger than that of a single model. Virtual ensemble that enables generating an ensemble for computing uncertainty using only a single trained GBDT model, since a GBDT model is itself an ensemble of trees. Virtual ensemble employs "truncated” sub-models of a single GBDT model as elements of an ensemble. As pointed in \cite{malinin2021uncertainty}, virtual ensemble can be obtained using any already constructed
GBDT model. From this method, model based uncertainties for the data instances can be obtained for both classification and regression tasks. 

\subsubsection{IBUG}
Another very recent work \cite{ibug}, proposed  \emph{Instance-Based Uncertainty estimation for Gradient-boosted regression trees (GBRT)} (\emph{IBUG}) for uncertainty estimation for regression task. IBUG can extend any GBRT point predictor to produce probabilistic
predictions. IBUG computes a non-parametric distribution around a prediction
using the k-nearest training instances, where distance is measured with a tree-ensemble kernel. Moreover, this method is applicable to any GBDT based regression models and not limited to CatBoost. 

For more details regarding the construction of virtual ensembles and IBUG, the readers are suggested to refer to \cite{malinin2021uncertainty} and \cite{ibug} repectively. 

\subsection{Cost Effective Active Learning}
Recently, a new approach called Cost-Effective Active Learning (CEAL) made AL even more efficient \cite{wang2016cost}. However, similar to the current trends in AL, this was only explored for deep neural networks and for the classification of image data. The central idea in CEAL is to utilize the datapoints from the unlabelled dataset and assign pseudo-labels for those datapoints which the model is highly confident in its prediction (based on entropy). These pseudo-labeled datapoints are then added to the next iterations of AL training. This is cost-effective since the labels are assigned by the model and not by human. However, in certain cases, the assigned pseudo-labels could be erroneous and hence the pseudo-labeled dataset could be noisy. To overcome this drawback of CEAL, we propose an additional filtering step for the classification task. Additionally, we propose a cost-effective active learning methodology for regression task as well. 

\section{Methods}

\subsection{AL using Model Uncertainty}

The central part in any AL framework is the sampling or query strategy. For the classification task, instead of employing the uncertainty sampling with entropy as the acquisition function, in this work, we propose the use of model uncertainty computed using virtual ensembles \cite{malinin2021uncertainty} as a drop-in replacement of entropy. The preliminary step as the case with generic AL frame is the initial training of the CatBoost model $h$ on $D_L$. The next step is to select a subset ($M$) of instances from the unlabelled set $D_U$ and query the labels for them for the next iteration of AL. For this, we first perform inference from the trained model $h$ and compute the uncertainties in the predictions for each of the instances $x_j$ in $D_U$ using the virtual ensembles \cite{malinin2021uncertainty}. Thus the utility score $s$ for a datapoint $x_j$ in $D_U$ is computed based on
\begin{equation}
    s(h,x_j) =  u_{virtual\ ensembles}(x_j)
\label{eq:ve}
\end{equation}
, where $u_{virtual\ ensembles}$ is the model uncertainty for $h$ computed using virtual ensembles \cite{malinin2021uncertainty}. We then rank the datapoints based on the highest model-uncertainties and select the top-$M$ samples for labelling. This process is then repeated until a desired performance is reached or a budget is exhausted. 

Apart from computing model-uncertainty based on virtual ensembles, we also leverage the individual trees of GBDT since GBDT itself is ensemble of trees. As pointed out in \cite{malinin2021uncertainty}, it should noted that in contrast to random forests
formed by independent trees \cite{Shaker_2020}, the sequential nature of GBDT models
implies that all trees are dependent and individual trees cannot be considered as separate models. However, for the task of active learning, instead of the actual value of model-uncertainty, we are mainly interested in ranking of the datapoints based on the model-uncertainty. For this sake, even a notion of relative uncertainty suffices. 

Consider a binary classification task for which the CatBoost model outputs the probabilities for the positive class $p(\vec{y}=1 \given \vec{x}; h)$. We can compute predictions for all the individual trees in the model $h$, i.e., $p_k = 
 p(y_j=1 \given x_j; h_k)$, $k = 1, \ldots, T$, where $T$ is the total number of individual trees in the CatBoost model $h$.
Similar to MC dropout \cite{gal2016dropout}, the standard deviation of the predictions can then be employed as the acquisition function: 
\begin{equation}
    s(x_j) = \sqrt{\frac{1}{T - 1} \sum_{k = 1}^T (p_k - \bar{p})^2},\ \bar{p} = \frac{1}{T} \sum_{k = 1}^T p_k
\label{eq:cbe}
\end{equation}
 The above method can be employed for multiclass classification as well.

For regression tasks, we again propose the use of model based uncertainty sampling as the query strategy. Again, virtual ensembles \cite{malinin2021uncertainty} can directly be applied for regression tasks (equation \ref{eq:ve}). Similar to the classification task, the individual trees of CatBoost can again be utilized for estimating uncertainty. This is similar to MCDUE \cite{tsymbalov2018dropout} 
 (equation \ref{eq:mcdue}), wherein instead of $T$ stochastic forward passes based on dropout, we obtain predictions from $T$ individual trees of CatBoost regressor i.e., $y_k = \hat{h}_{k}(x_j)$,  $k = 1, \ldots, T$. The standard deviation (which is the CatBoost model uncertainty) of the predictions is then employed as the acquisition function: 
\begin{equation}
    s(x_j) = \sqrt{\frac{1}{T - 1} \sum_{k = 1}^T (y_k - \bar{y})^2},\ \bar{y} = \frac{1}{T} \sum_{k = 1}^T y_k
\label{eq:cbr}
\end{equation}
The uncertainties computed based on the standard deviations from equations \ref{eq:cbe} and \ref{eq:cbr} is referred to as uncertainty from \emph{staged predictions}. Thus, the utility score $s$ for a datapoint $x_j$ in $D_U$ can then be written in short:
\begin{equation}
    s(h,x_j) =  u_{staged\ predictions}(x_j)
\label{eq:st}
\end{equation}
Apart from the above methods, we propose the use of additional method for computing model-uncertainties for CatBoost regressor based on the method IBUG proposed in \cite{ibug}. Thus the utility score $s$ for a datapoint $x_j$ in $D_U$ is computed based on
\begin{equation}
    s(h,x_j) =  u_{ibug}(x_j)
\label{eq:ibug} 
\end{equation}
%------------ Start of Figure Classification AL AUC

\begin{figure*}[!htbp]
    \begin{subfigure}{0.32\textwidth}
        \includegraphics[width=\linewidth]{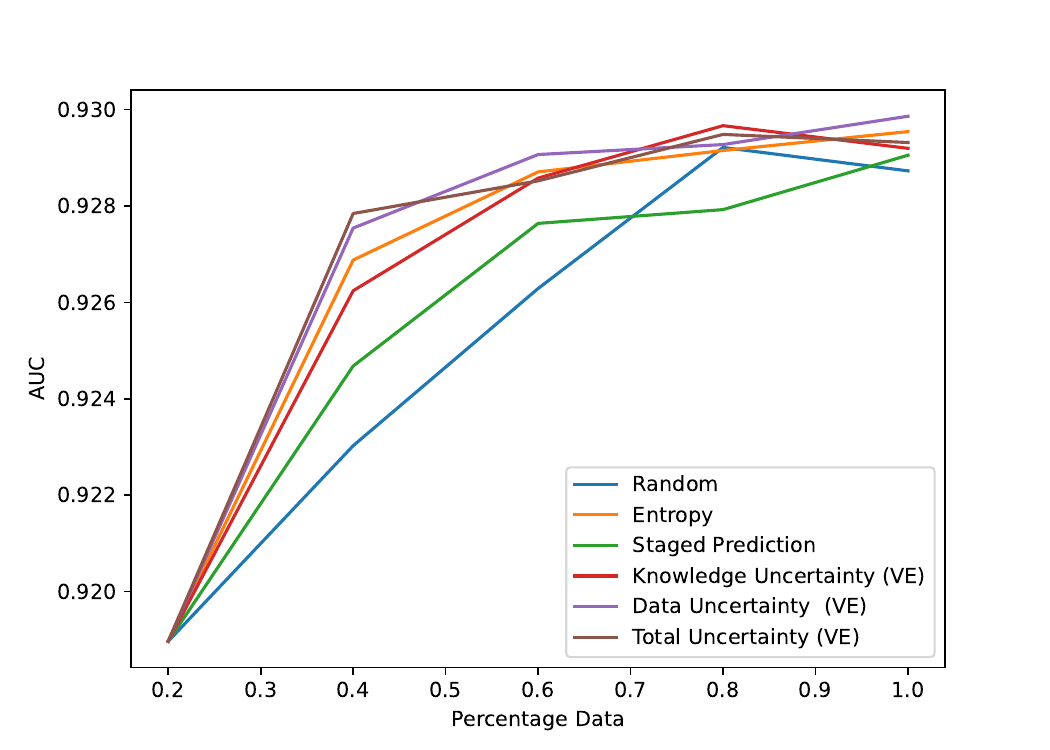}
        \caption{Adult Dataset}
    \label{subfig:Adult_al_comp}
    \end{subfigure}
    \hfill
    \begin{subfigure}{0.32\textwidth}
        \includegraphics[width=\linewidth]{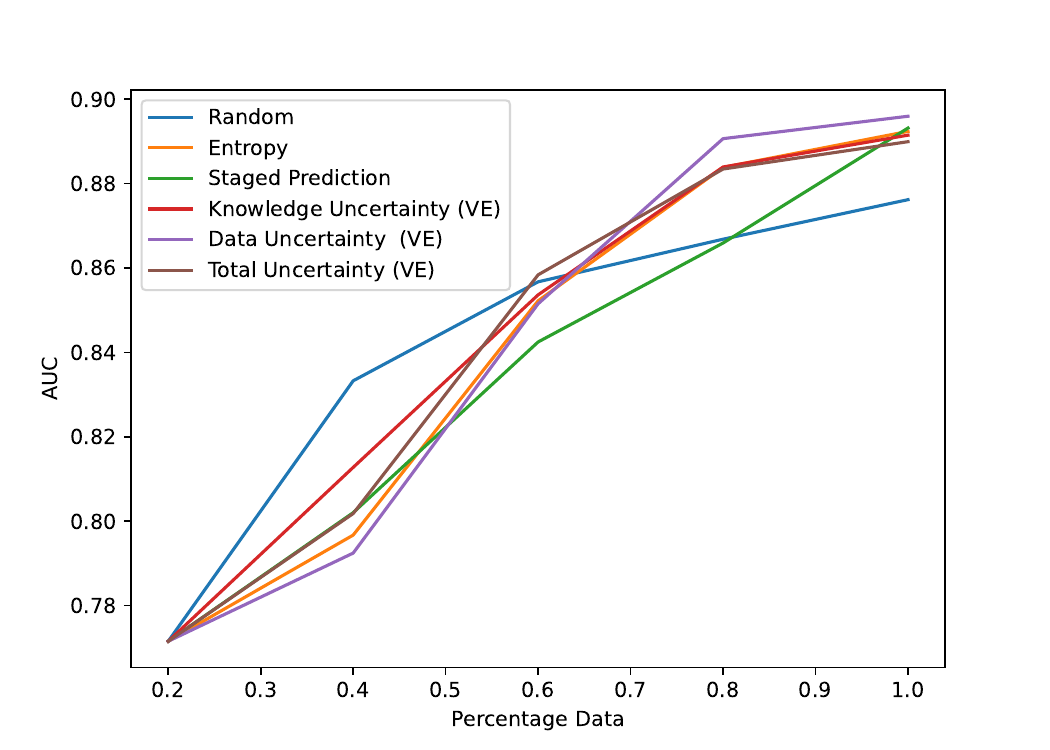}
        \caption{Amazon Dataset}
        \label{subfig:Amazon_al_comp}
    \end{subfigure}
    \hfill
    \begin{subfigure}{0.32\textwidth}
        \includegraphics[width=\linewidth]{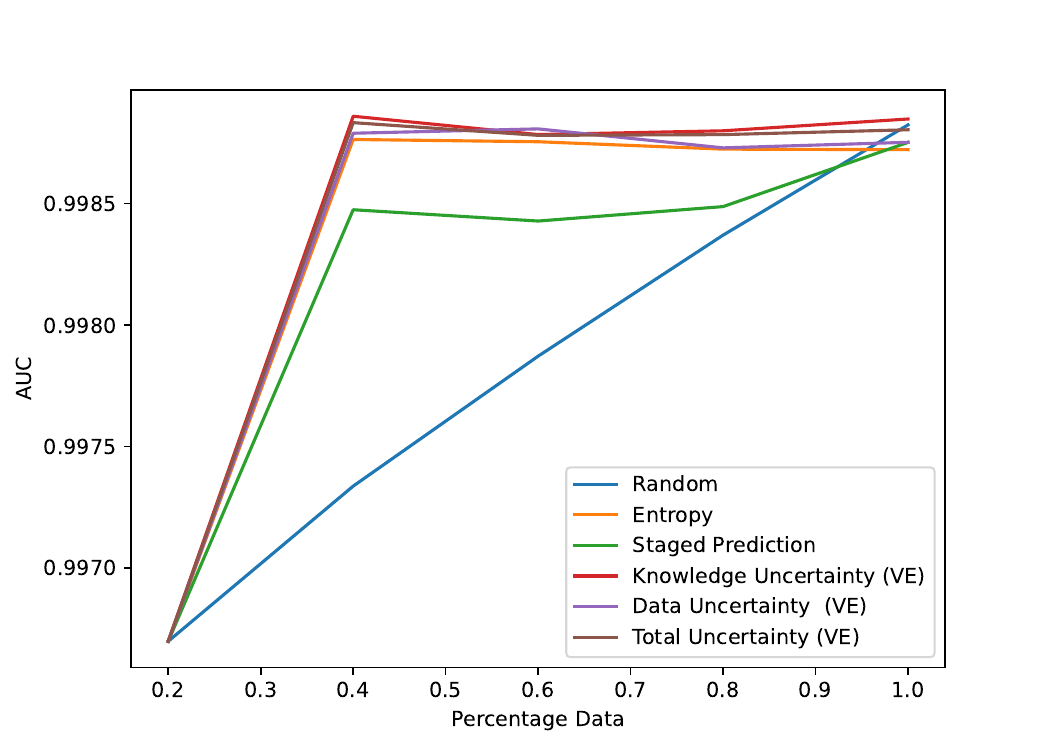}
        \caption{Thyroid Dataset}
        \label{subfig:Thyroid_al_comp}
    \end{subfigure}
    \hfill
    \caption{Comparison of various sampling strategies for classification task using AUC metric}
    \label{fig:AUC_no_CEAL}
\end{figure*}

%------------ Start of Figure Regression AL MSE

\begin{figure*}[!htbp]
    \begin{subfigure}{0.32\textwidth}
        \includegraphics[width=\linewidth]{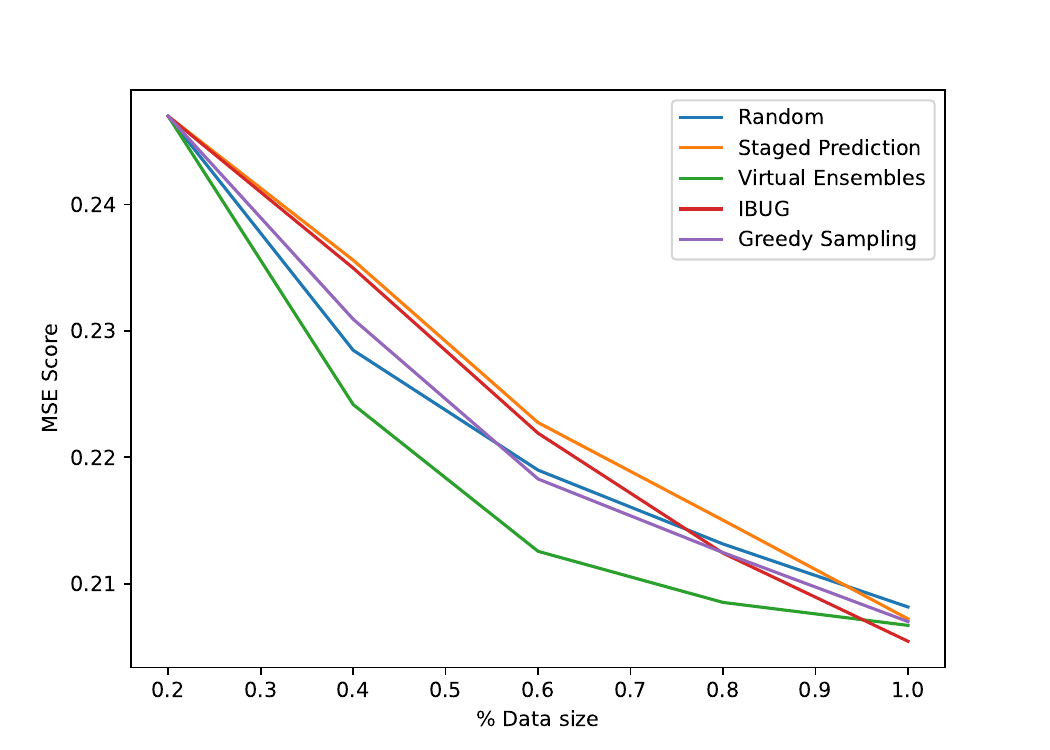}
        \caption{California Housing Dataset}
    \label{subfig:cal_no_ceal}
    \end{subfigure}
    \hfill
    \begin{subfigure}{0.32\textwidth}
        \includegraphics[width=\linewidth]{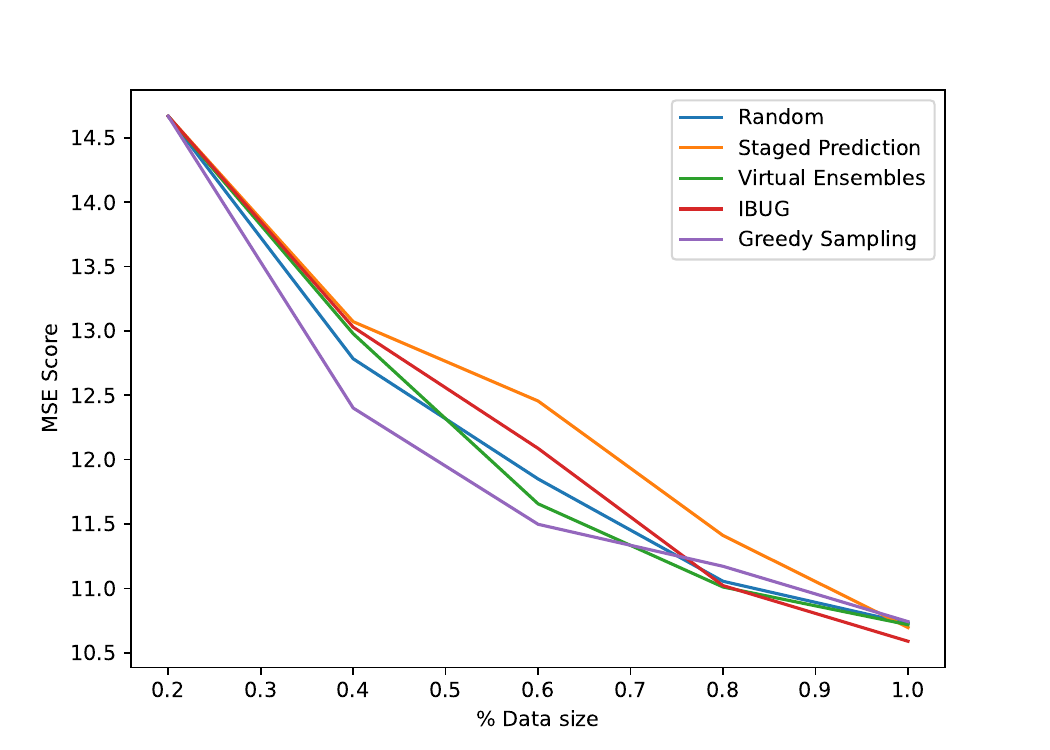}
        \caption{Energy Dataset}
        \label{subfig:energy_no_ceal}
    \end{subfigure}
    \hfill
    \begin{subfigure}{0.32\textwidth}
        \includegraphics[width=\linewidth]{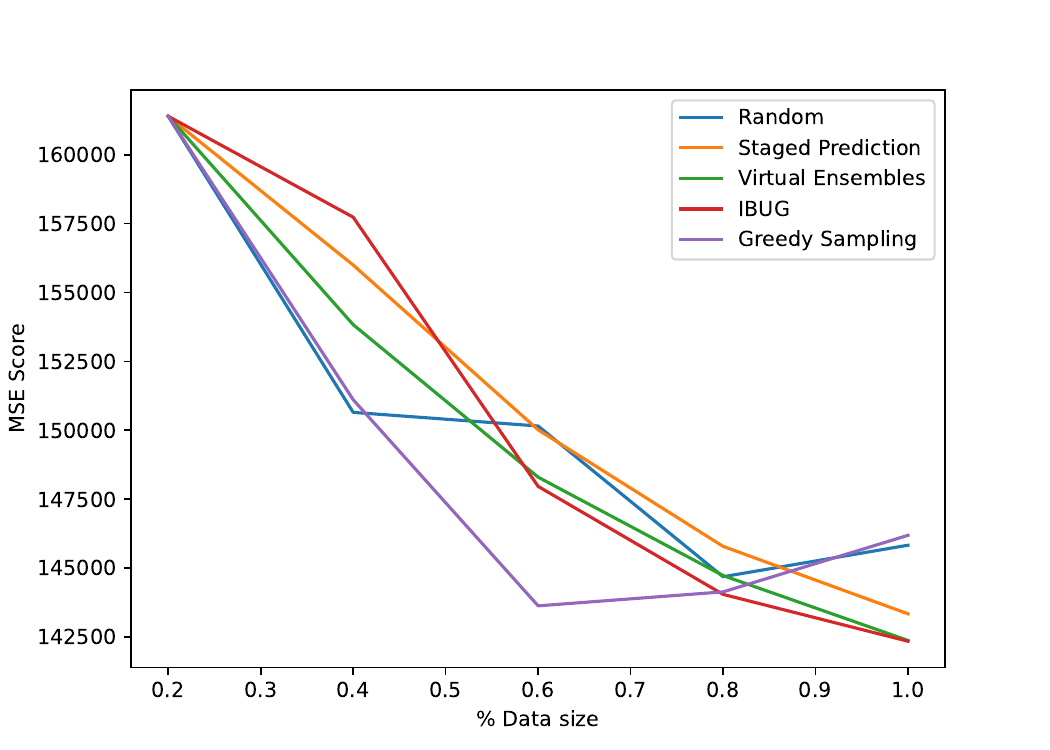}
        \caption{Communities and Crime Dataset}
        \label{subfig:comm_no_ceal}
    \end{subfigure}
    \hfill
    \caption{Comparison of various sampling strategies for regression task using MSE metric}
    \label{fig:MSE_no_CEAL}
\end{figure*}

\subsection{CEAL}
\subsubsection{Classification Task}
The proposed utilization of model uncertainty has an added benefit that can be useful in overcoming the drawbacks of CEAL \cite{wang2016cost}. A classification model tends to be overconfident in it's predictions and especially so when trained with less data. Assigning pseudo-labels by calculating the entropy alone would lead to erroneous labelling. Hence, we propose the following steps for obtaining accurate pseudo-labels for CEAL:
\begin{enumerate}
    \item Set an entropy threshold $\delta_{entropy}$. This value is set based on the difficulty in prediction. To get a very high confidence datapoints, entropy threshold is set to a small value. 
    \item In the first stage of selection or filtering, select those datapoints as the candidates for high-confidence set, whose entropy values are less than the entropy threshold $\delta_{entropy}$. Lesser the entropy, higher the confidence.
    \item Set an additional threshold for \emph{model uncertainty} 
    $\delta_{uncertainty}$. To get high confidence datapoints, the threshold is set to a small value. This model uncertainty can be computed based on any of the methods such as virtual ensembles (equation \ref{eq:ve}) or staged predictions (equation \ref{eq:st}).
    \item In the second stage of selection or filtering, select those datapoints from the already filtered high-confidence points in step 2, whose model uncertainty  values are less than the model uncertainty threshold $\delta_{uncertainty}$. Lesser the model uncertainty, higher the confidence.
    \item The resultant high confidence datapoints obtained from the two stage filtering are then labeled based on the outputs of the classifier.
\end{enumerate}
Thus the pseudo-label is defined as as:
\begin{equation}\label{equ:pseudo_class}
\begin{gathered}
\lambda^* = \arg \max_{\lambda} p( y_j=\lambda \vert x_j; \mathcal{\theta}), \\
y_j = \left\{
\begin{array}{c}
 \\
  \lambda^*, \\
 \\
 \\
0,%
\end{array}
\begin{array}{c}
entropy_j < \delta_{entropy_j}, \\ 
\\
model\_uncertainty_j < \delta_{uncertainty_j} \\
\\
\text{otherwise.}
\end{array}
\right.
\end{gathered}	
\end{equation}

\subsubsection{Regression Task}
The use of model uncertainty for regression also helps us in proposing a CEAL methodology for regression task not proposed before.
\begin{enumerate}
    \item Set a threshold for \emph{model uncertainty} 
    $\delta_{uncertainty}$. To get high confidence datapoints, the threshold is set to a small value. This model uncertainty can be computed based on any of the methods such as virtual ensembles (equation \ref{eq:ve}) or staged predictions (equation \ref{eq:st}) or IBUG ( equation \ref{eq:ibug}).
    \item Select those datapoints from the unlabelled set $D_U$ whose model uncertainty  values are less than model uncertainty threshold $\delta_{uncertainty}$. Lesser the model uncertainty, higher the confidence.
    \item The resultant high confidence datapoints obtained are then labeled based on the outputs obtained from the regressor.
\end{enumerate}
Thus the pseudo-label is defined as as:
\begin{equation}\label{equ:pseudo_regress}
\begin{gathered}
y^* = h(x_j) \\
y_j = \left\{
\begin{array}{c}
  y^*, \\
0,%
\end{array}
\begin{array}{c}
model\_uncertainty_j < \delta_{uncertainty_j} \\
\text{otherwise.}
\end{array}
\right.
\end{gathered}	
\end{equation}

% %%%%%%%%%%%%%%% Algo
% % \begin{figure}
% % \vspace{-5em}
% % \begin{spacing}{0.8}
% \begin{algorithm}[!tp]
% \caption{ Learning Algorithm of proposed AL framework}
% \label{alg:HybridCEAL}
% \begin{algorithmic}[1]
% \REQUIRE ~~\\
% Unlabeled samples ${D}^U$, initially labeled samples ${D}^L$, uncertain samples selection size $m$, outlier probability threshold $\delta_{p_{outlier}}$, OD model's confidence threshold $\delta_{C_{OD}}$,
% high confidence samples selection threshold for entropy $\delta_{entropy}$, threshold decay rate $dr$, threshold for inlier probability $\delta_{p_{inlier}}$, maximum iteration number $T$, fine-tuning interval $t$.
% \ENSURE ~~\\
% Model or model parameters $\mathcal{\theta}$.
% \STATE { Initialize $\mathcal{\theta}$ with ${D}^L$.}
% \WHILE{ { not reach maximum iteration $T$}}
% \STATE {Add $m$ uncertain samples into $D^L$ based on Eq.~(\ref{eq:en})}
% \STATE {Add $n$ diverse samples into $D^L$ based on OD model}
% \STATE {Obtain high confidence samples $D^H$ based on  Eq.~(\ref{equ:pseudo})}
% \STATE {In every $t$ iterations:}\begin{itemize}\setlength{\itemindent}{1em}
% \item {Update  $\mathcal{\theta}$}
% \item {Update $\delta_{entropy}$ (if required) 
% according to the method described in \cite{wang2016cost}}
% \end{itemize}
% \ENDWHILE
% \RETURN $\mathcal{\theta}$
% \end{algorithmic}
% \end{algorithm}
% % \end{spacing}
% % \vspace{-5em}
% % \end{figure}
% %%%%%%%%%%%%%%%%%%%%%%%%%
\section{Experiments}
For the experiments, we employ CatBoost  \cite{prokhorenkova2018catboost} classifier and regressor as our base models for the respective tasks. For classification task, we perform experiments on three datasets namely, \textit{Adult} dataset \cite{misc_adult_2} ,  \textit{Amazon} dataset \cite{amazon} and and \textit{Thyroid} dataset \cite{misc_thyroid_disease_102}. The first two are binary classification datasets and the third one is a multiclass classification dataset. For regression task, we again employ three datasets namely, \textit{California Housing} dataset \cite{california_housing}, 
\textit{Energy} dataset \cite{energy} and \textit{Communities and Crime} dataset \cite{communities_and_crime}. For all the datasets, we split 80\%-20\% for training and testing, respectively. 

We perform two sets of experiments for each of the classification and regression tasks. For the classification task, we compare the model uncertainty based sampling with the traditionally employed entropy based uncertainty sampling as well as random sampling, in first set of experiments. Similarly, for the regression task, we compare the model uncertainty based sampling with the previous state-of-the-art greedy sampling method proposed in \cite{wu2019active} as well as random sampling. 
%------------ Start of Figure Classification CEAL AUC

\begin{figure*}[!htbp]
    \begin{subfigure}{0.32\textwidth}
        \includegraphics[width=\linewidth]{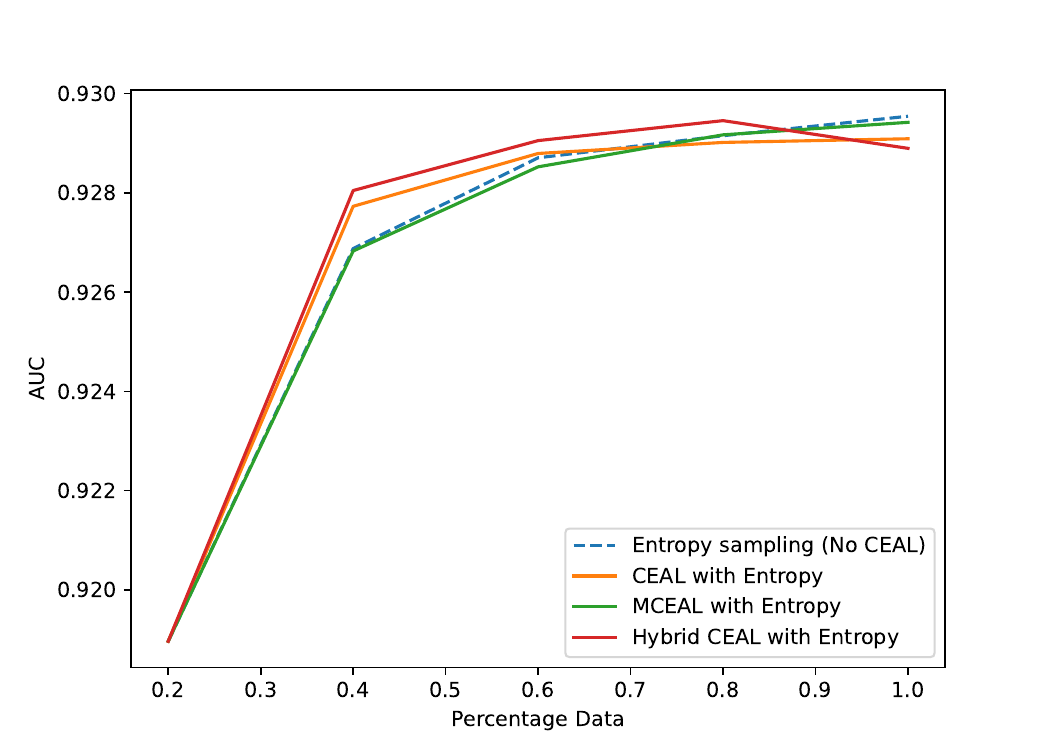}
        \caption{Adult Dataset}
    \label{subfig:Adult_ceal_comp_es}
    \end{subfigure}
    \hfill
    \begin{subfigure}{0.32\textwidth}
        \includegraphics[width=\linewidth]{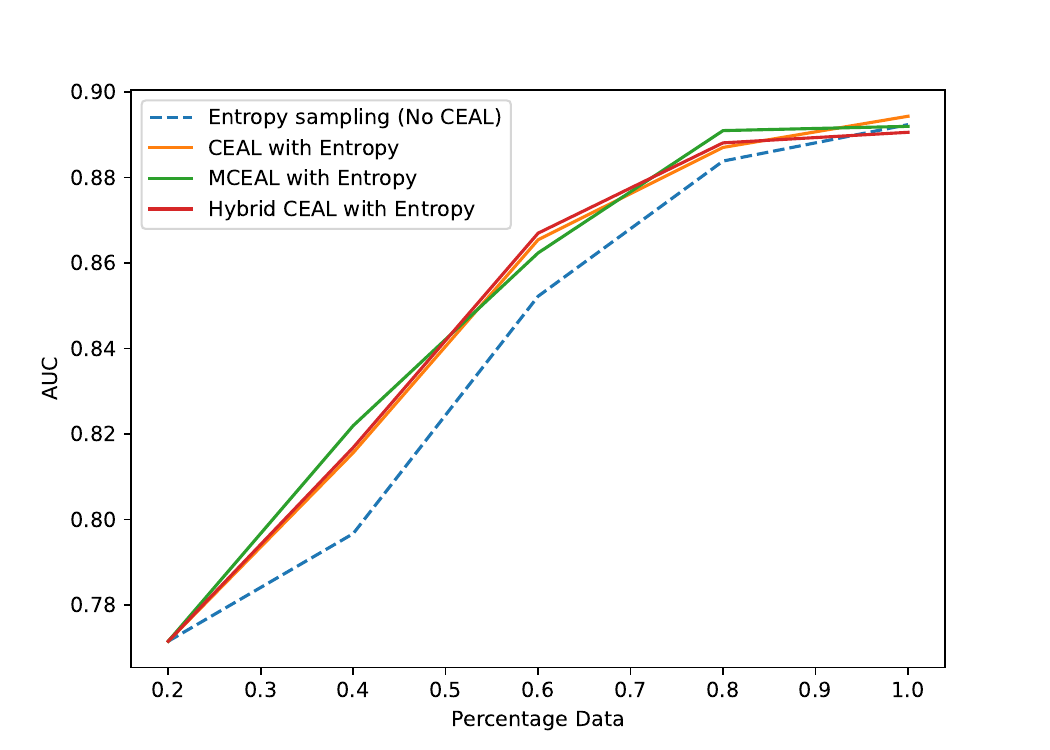}
        \caption{Amazon Dataset}
        \label{subfig:Amazon_ceal_comp_es}
    \end{subfigure}
    \hfill
    \begin{subfigure}{0.32\textwidth}
        \includegraphics[width=\linewidth]{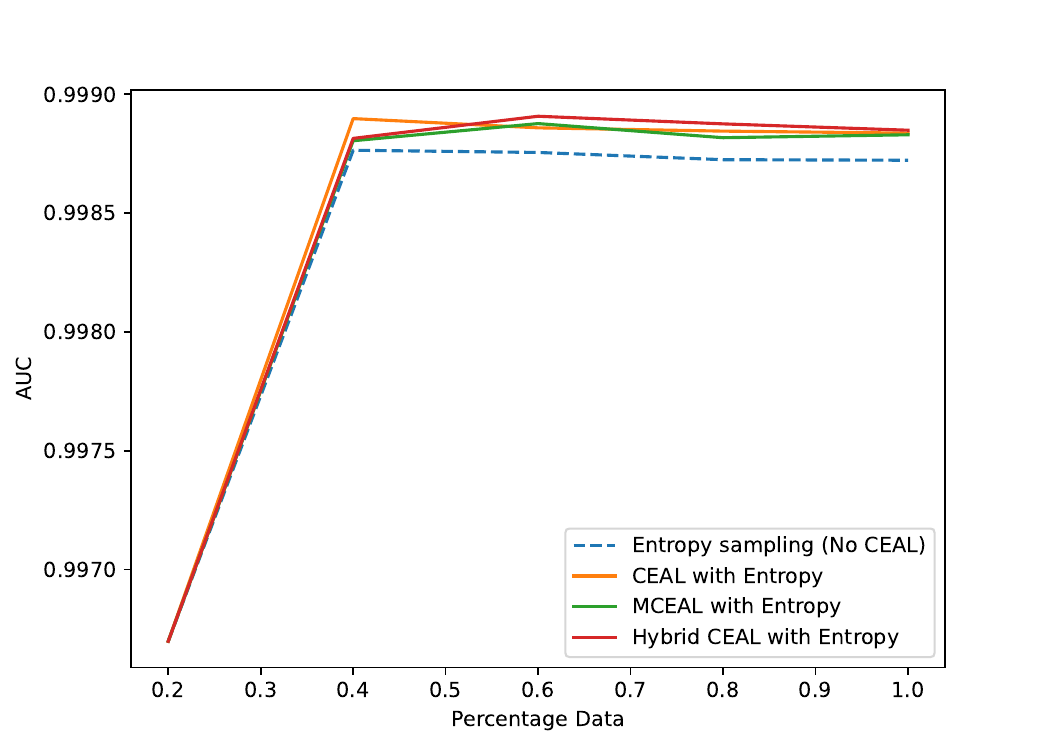}
        \caption{Thyroid Dataset}
        \label{subfig:Thyroid_ceal_comp_es}
    \end{subfigure}
    \hfill
    \newline
        \begin{subfigure}{0.32\textwidth}
        \includegraphics[width=\linewidth]{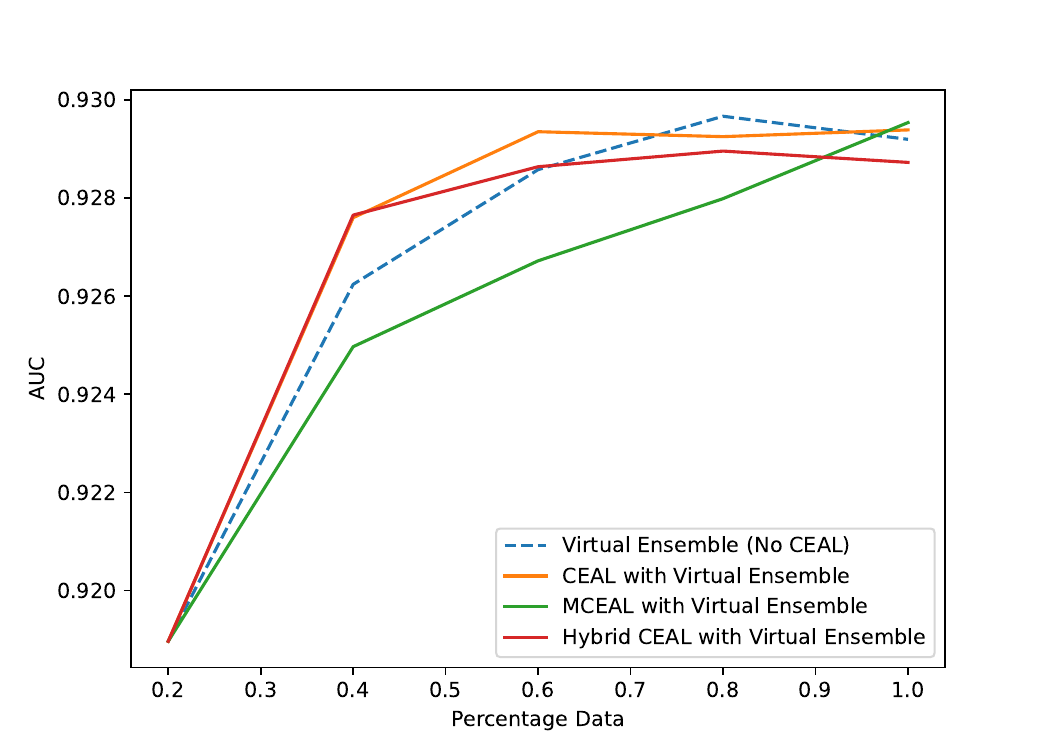}
        \caption{Adult Dataset}
    \label{subfig:Adult_ceal_comp_ve}
    \end{subfigure}
    \hfill
    \begin{subfigure}{0.32\textwidth}
        \includegraphics[width=\linewidth]{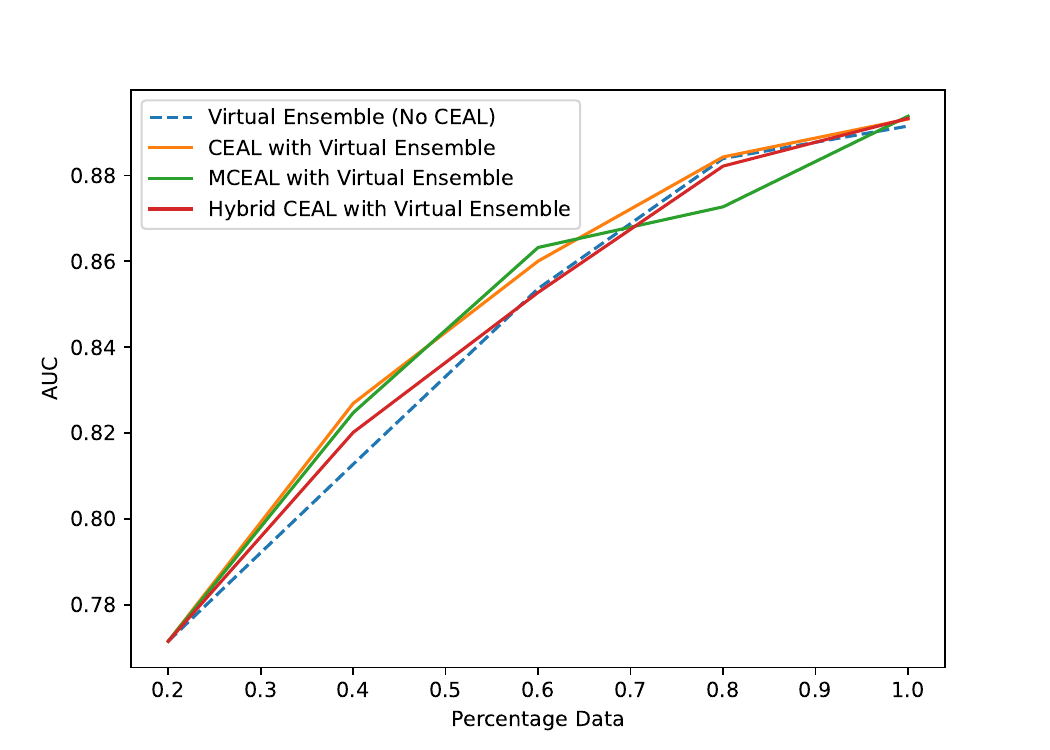}
        \caption{Amazon Dataset}
        \label{subfig:Amazon_ceal_comp_ve}
    \end{subfigure}
    \hfill
    \begin{subfigure}{0.32\textwidth}
        \includegraphics[width=\linewidth]{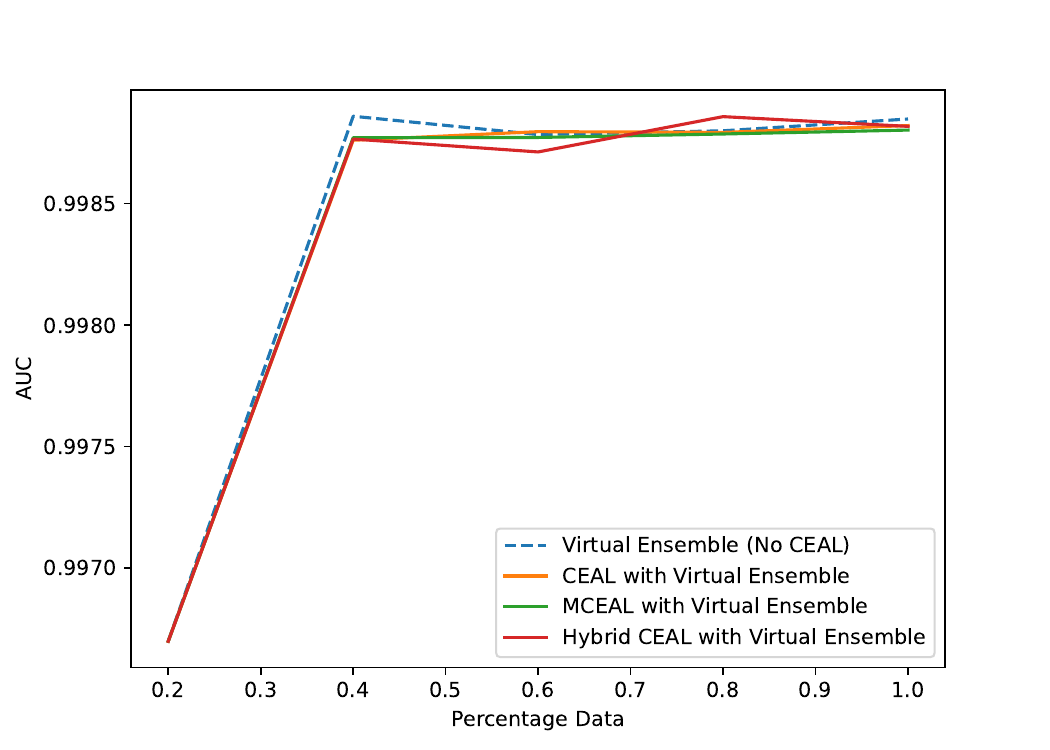}
        \caption{Thyroid Dataset}
        \label{subfig:Thyroid_ceal_comp_ve}
    \end{subfigure}
    \hfill
    \caption{Comparison of different variants of CEAL for Classification task. The top row contains results for experiments using entropy sampling as the query strategy and the bottom row contains results for experiments using model based uncertainty sampling using virtual ensembles as the query strategy }
    \label{fig:AUC_CEAL_ES_vs}
\end{figure*}
For the second set of experiments, we focus on different cost-effective active learning (CEAL) regimes. For the classification task, we compare with traditional CEAL proposed in \cite{wang2016cost} along with the proposed modifications to CEAL for high confidence data sampling. For the regression task, we compare the AL with and without the addition of pseudo-labels (i.e, CEAL vs No-CEAL). Thus, our experiments are conducted in a way to address the following questions:
\begin{itemize}
    \item How does model uncertainty sampling compare with traditional sampling methods for classification and regression tasks?
    \item How does the proposed modifications to CEAL compare with the traditional CEAL for classification task?
    \item How effective is CEAL or the addition of pseudo-labels for the regression task?
\end{itemize}

\section{Results and Discussion}
For comparing the results of various experiments, we employ accuracy and area under curve (AUC) as the metrics for classification tasks, and employ mean squared error (MSE) and coefficient of determination ($R^2$) regression score (R2 Score) as the metrics for regression tasks. For all the AL experiments, we perform the initial training with only $20\%$ of the training set and thereby incrementally add $20\%$ at each iteration based on the query strategy employed.

%%%%%%%%%%%%%%%%%%%%%%%%%%%%%%%%%%%%%%%%%%%%%%%%%%%%%%%%%%%%
\subsection{Comparison of different sampling strategies}

\subsubsection{Classification}

The plots in the figure Fig. \ref{fig:AUC_no_CEAL} contains results obtained for the three datasets employed for the classification task. We employed the following sampling strategies in our experiments on classification datasets.
\begin{itemize}
    \item \textit{Random} sampling as the baseline sampling method wherein we randomly sample a fixed percentage of labelled data at every AL iteration
    \item  \textit{Entropy} sampling wherein at every AL iteration, we sample a fixed  percentage of labelled datapoints based on entropy as the acquisition function or score function based on equation \ref{entropy}.
    \item Model uncertainty sampling using \textit{staged prediction}, wherein we sample the uncertain datapoints based on the scores obtained from equations \ref{eq:st}, \ref{eq:cbe}.
    \item Model uncertainty sampling based on \textit{virtual ensembles (VE)} proposed in the work \cite{malinin2021uncertainty}. For VE based uncertainty sampling, we compute three different types of uncertainty as proposed in \cite{malinin2021uncertainty}.
    \begin{itemize}
        \item Knowledge Uncertainty
        \item Data Uncertainty
        \item Total Uncertainty
    \end{itemize}
\end{itemize}

From the plots of the figure Fig. \ref{fig:AUC_no_CEAL}, it can indeed be seen that on real-world dataset it is very hard to beat random sampling. This has been an area of interest recently to explore the practical challenges of AL deployment \cite{lowell2019practical, kagy2019practical}, where the authors perform detailed analysis of the passive vs active sampling strategy. For the \textit{Thyroid} dataset, it can be seen that active sampling methods clearly perform better than random sampling. Similar trends can be seen in Adult dataset when the availability of data is low. For Amazon dataset, random sampling performing better than active sampling methods in low data availability regime is an anomaly. When comparing different active uncertainty sampling methods, it can be seen that apart from model-based uncertainty sampling methods perform similar to entropy based uncertainty sampling, with the exception of model uncertainty computed from staged predictions. Uncertainty computed from  Virtual ensembles tend to perform better than entropy in most cases. Although, knowledge uncertainty (computed through virtual ensembles of CatBoost) theoretically represents the gaps in the model knowledge, in most cases, it performs slightly lesser when compared to total uncertainty, which is the combination of both the knowledge uncertainty and the expected data uncertainty \cite{malinin2021uncertainty}.
%------------ Start of Figure Regression CEAL MSE

\begin{figure*}[!htbp]
    \begin{subfigure}{0.32\textwidth}
        \includegraphics[width=\linewidth]{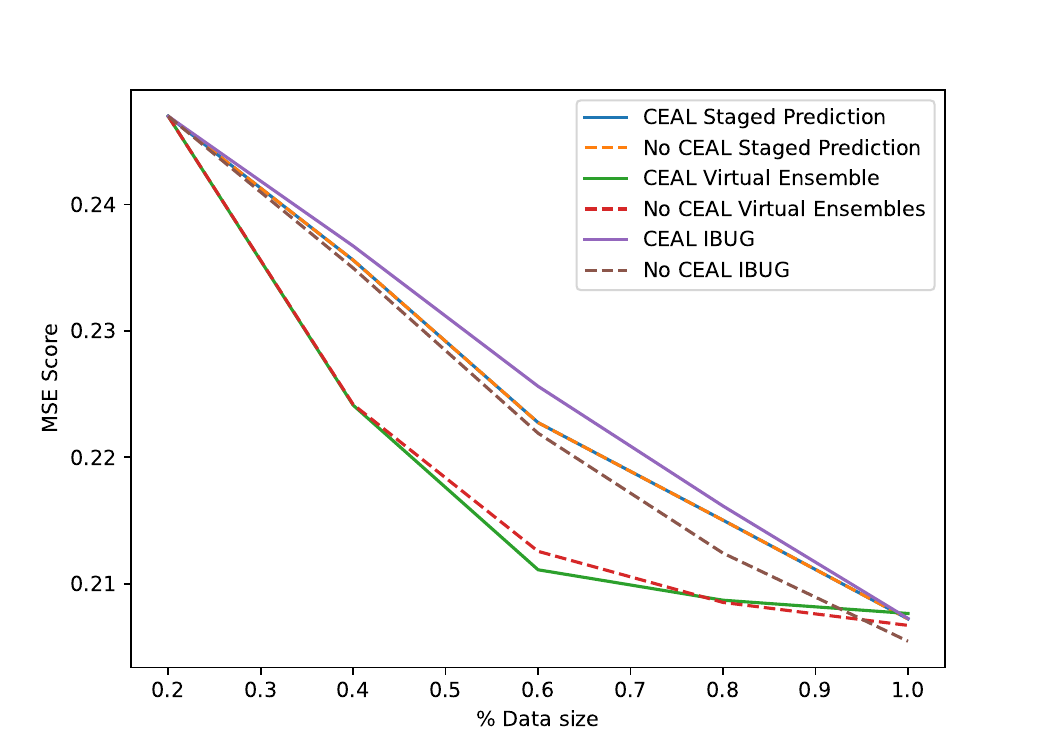}
        \caption{California Housing Dataset}
    \label{subfig:cal_ceal}
    \end{subfigure}
    \hfill
    \begin{subfigure}{0.32\textwidth}
        \includegraphics[width=\linewidth]{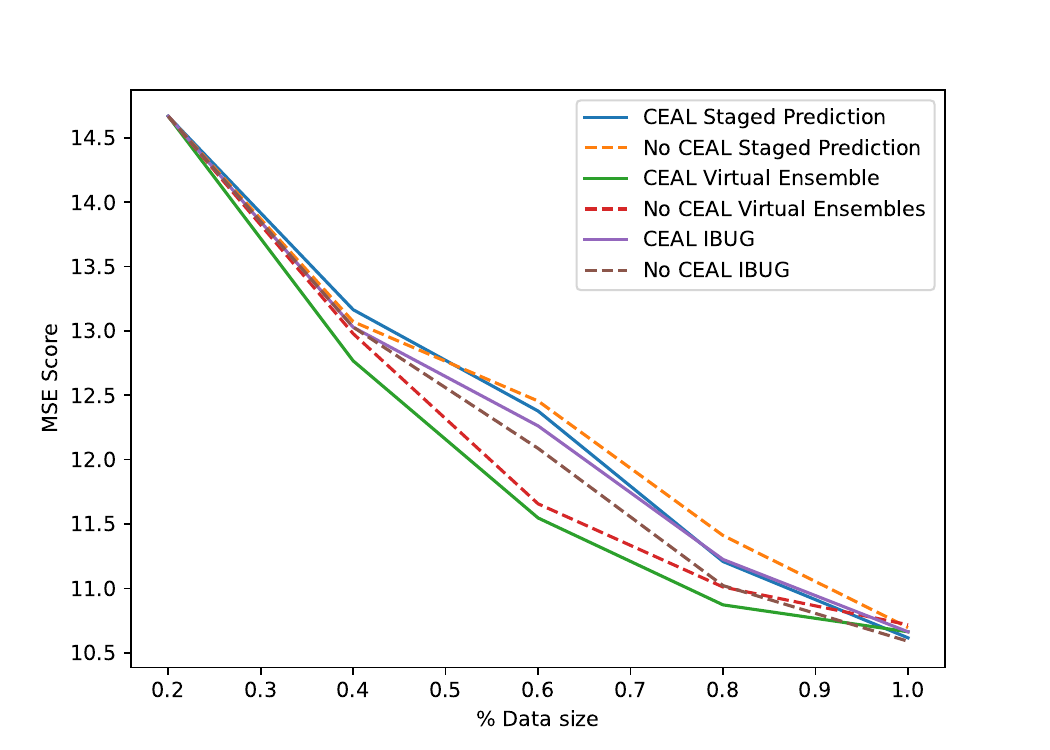}
        \caption{Energy Dataset}
        \label{subfig:energy_ceal}
    \end{subfigure}
    \hfill
    \begin{subfigure}{0.32\textwidth}
        \includegraphics[width=\linewidth]{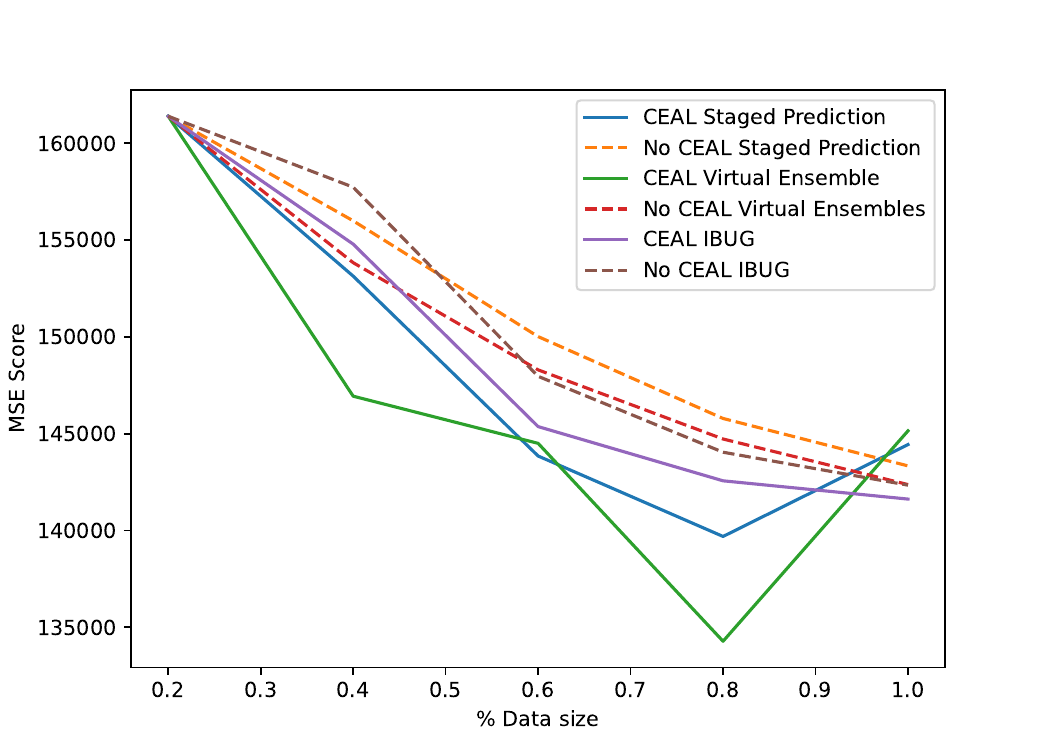}
        \caption{Communities and Crime Dataset}
        \label{subfig:comm_ceal}
    \end{subfigure}
    \hfill
    \caption{Comparison of various CEAL variants for regression task using MSE metric. The dotted line represent absence of pseudo-label sampling (or CEAL), while solid line indicates the presence of CEAL for different model based uncertainty sampling strategies.}
    \label{fig:MSE_CEAL}
\end{figure*}
%%%%%%%%%%%%%%%%%%%%%%%%%%
\subsubsection{Regression}

The plots in the figure Fig. \ref{fig:MSE_no_CEAL} contains results obtained for the three datasets employed for the regression task. We employed the following sampling strategies in our experiments for the regression task.
\begin{itemize}
    \item \textit{Random} Sampling
    \item \textit{Greedy sampling} proposed in the work \cite{wu2019active} 
    \item Model uncertainty sampling using \textit{staged prediction}, wherein we sample the uncertain datapoints based on the scores obtained from equations \ref{eq:st} and \ref{eq:cbr}.
    \item Model uncertainty sampling based on \textit{virtual ensembles (VE)} proposed in the work \cite{malinin2021uncertainty}.
    \item Model uncertainty sampling based on \textit{IBUG} proposed in the work \cite{ibug}.
\end{itemize}

Similar to the trends seen in classification task, it can be seen from figure Fig. \ref{fig:MSE_no_CEAL} that it can be challenging for active sampling techniques to outperform passive techniques like random sampling even for regression task. It should be noted that the greedy sampling method is also a passive sampling method. However, greedy sampling has an exponential time-complexity making it unviable for practical use cases and more so when there is a need to sample large number of data instances. Amongst the model based uncertainty sampling methods, IBUG tends to converge to the least MSE score in the final AL iterations. For the California Housing dataset, model uncertainty sampling based on virtual ensembles is clearly the best performing method. Amongst the model based uncertainty sampling methods, virtual ensembles has the least time complexity while IBUG has the highest. However, IBUG is applicable to any boosted trees model and not limited to CatBoost which is not the case for other model based uncertainty estimation methods.

\subsection{Comparison of different CEAL variants}
\subsubsection{Classification}
For the second set of experiments, we examine different variants of CEAL based on the proposed modification. The resulting plots obtain for these set of experiments for classification task is shown in figure Fig. \ref{fig:AUC_CEAL_ES_vs}. Different variants of CEAL considered are as follows:
\begin{itemize}
    \item CEAL with entropy: This is the traditional CEAL methodology as described in \cite{wang2016cost}, wherein high confidence samples are selected based on entropy for pseudo-labelling and with entropy as the acquisition function for uncertainty sampling. 
    \item MCEAL with entropy: In contrast to traditional CEAL, we sample high confidence samples based on model's knowledge uncertainty obtained from virtual ensembles and using entropy as the acquisition function for sampling low-confidence datapoints for oracle annotation in uncertainty sampling.
    \item Hybrid CEAL with entropy: In this we employ a two stage filtering criteria, wherein we first sample high confidence datapoints based on entropy and further filter the obtained high-confidence datapoints based on model's knowledge uncertainty obtained from virtual ensemble. Thus, this technique involves sampling high-confidence points using both entropy and model uncertainty, while employing entropy based uncertainty for sampling low-confidence datapoints.    
\end{itemize}
In addition to the above, we perform the similar experiments by employing model uncertainty (computed using virtual ensembles) for sampling low-confidence datapoints. The three additional experiments are named as:
\begin{itemize}
    \item CEAL with Virtual Ensemble
    \item MCEAL with Virtual Ensemble
    \item Hybrid CEAL with Virtual Ensemble
\end{itemize}
The top and bottom rows of figure Fig. \ref{fig:AUC_CEAL_ES_vs} correspond to experiments with uncertainty sampling using entropy and virtual ensembles respectively. It can be seen that in most cases CEAL proves to be beneficial. CEAL involves proper tuning of the thresholding parameter for selection of high confidence points. Although hybrid CEAL performs better than other CEAL variants in most cases, it should be noted that it requires tuning of two threshold parameters- one for entropy and the other for model uncertainty.

\subsubsection{Regression}: For verifying the usefulness of the proposed novel cost-effective active learning for regression task, we performed experiments by employing pseudo-labels obtained from high confidence datapoints obtained from model uncertainty, the results of which are shown in figure Fig. \ref{fig:MSE_CEAL}. 
Apart from sampling low-confidence datapoints for querying labels from the oracle  based on model uncertainty, we also sample high-confidence datapoints from the respective methods employed for computing model uncertainty such as \textit{staged prediction}, \textit{virtual ensembles}, \textit{IBUG}. The figure Fig. \ref{fig:MSE_CEAL} shows comparison for different methods with and without CEAL. It can be clearly seen from the figures that all the model-uncertainty based sampling methods, it is advantageous have the high-confidence pseudo-labels aided cost effective active learning. CEAL performs either better or matches it's counterparts. CEAL employing virtual ensembles as method for computing model uncertainty performs the best in all the cases, especially when the amount of labelled data is less available. Thus, looking holistically  at all the set of experiments, sampling based on model uncertainty from virtual ensembles seems very promising for both classification and regressions.  

\section{Conclusion}
In this work, we proposed model uncertainty based active learning for boosted trees. To the best of our knowledge, this was the first work that explored a non-Bayesian approach for model uncertainty based sampling for active learning. From our experiments, we showed that model uncertainty can be used as a drop-in replacement of traditional uncertainty measures such as entropy. 
From our experiments, it can be seen that model uncertainty seems promising as the sampling strategy for both classification and regression tasks. Moreover, we proposed an improved CEAL methodology for classification tasks wherein we leveraged both entropy as well as model uncertainty. We also proposed a CEAL framework for regression tasks and experimentally showed the benefits. 
In future, we would like to explore model uncertainty using other recent boosted tree methods such as \emph{NGBoost}
\cite{duan2020ngboost}. 
Another line of future work could be to further explore CEAL methods to reduce the dependence on hyperparameter tuning of thresholds employed for high-confidence data selection. 
\bibliographystyle{ACM-Reference-Format}
\bibliography{biblio_1}


\begin{thebibliography}{49}

%%% the Web address.

\ifx \showCODEN    \undefined \def \showCODEN     #1{\unskip}     \fi
\ifx \showDOI      \undefined \def \showDOI       #1{#1}\fi
\ifx \showISBNx    \undefined \def \showISBNx     #1{\unskip}     \fi
\ifx \showISBNxiii \undefined \def \showISBNxiii  #1{\unskip}     \fi
\ifx \showISSN     \undefined \def \showISSN      #1{\unskip}     \fi
\ifx \showLCCN     \undefined \def \showLCCN      #1{\unskip}     \fi
\ifx \shownote     \undefined \def \shownote      #1{#1}          \fi
\ifx \showarticletitle \undefined \def \showarticletitle #1{#1}   \fi
\ifx \showURL      \undefined \def \showURL       {\relax}        \fi
% invisible to TeX
\providecommand\bibfield[2]{#2}
\providecommand\bibinfo[2]{#2}
\providecommand\natexlab[1]{#1}
\providecommand\showeprint[2][]{arXiv:#2}

\bibitem[Arik and Pfister(2021)]%
        {arik2021tabnet}
\bibfield{author}{\bibinfo{person}{Sercan~{\"O} Arik} {and} \bibinfo{person}{Tomas Pfister}.} \bibinfo{year}{2021}\natexlab{}.
\newblock \showarticletitle{Tabnet: Attentive interpretable tabular learning}. In \bibinfo{booktitle}{\emph{Proceedings of the AAAI conference on artificial intelligence}}, Vol.~\bibinfo{volume}{35}. \bibinfo{pages}{6679--6687}.
\newblock


\bibitem[Becker and Kohavi(1996)]%
        {misc_adult_2}
\bibfield{author}{\bibinfo{person}{Barry Becker} {and} \bibinfo{person}{Ronny Kohavi}.} \bibinfo{year}{1996}\natexlab{}.
\newblock \bibinfo{title}{{Adult}}.
\newblock \bibinfo{howpublished}{UCI Machine Learning Repository}.
\newblock
\newblock
\shownote{{DOI}: https://doi.org/10.24432/C5XW20}.


\bibitem[Bilgic et~al\mbox{.}(2010)]%
        {bilgic2010active}
\bibfield{author}{\bibinfo{person}{Mustafa Bilgic}, \bibinfo{person}{Lilyana Mihalkova}, {and} \bibinfo{person}{Lise Getoor}.} \bibinfo{year}{2010}\natexlab{}.
\newblock \showarticletitle{Active learning for networked data}. In \bibinfo{booktitle}{\emph{Proceedings of the 27th international conference on machine learning (ICML-10)}}. \bibinfo{pages}{79--86}.
\newblock


\bibitem[Bittling et~al\mbox{.}(2017)]%
        {amazon}
\bibfield{author}{\bibinfo{person}{Mayer Bittling}, \bibinfo{person}{Kaggle}, {and} \bibinfo{person}{Amazon}.} \bibinfo{year}{2017}\natexlab{}.
\newblock \bibinfo{title}{{Amazon dataset}}.
\newblock
\newblock
\urldef\tempurl%
\url{https://www.kaggle.com/bittlingmayer/amazonreviews}
\showURL{%
\tempurl}


\bibitem[Brophy and Lowd(2022)]%
        {ibug}
\bibfield{author}{\bibinfo{person}{Jonathan Brophy} {and} \bibinfo{person}{Daniel Lowd}.} \bibinfo{year}{2022}\natexlab{}.
\newblock \showarticletitle{Instance-Based Uncertainty Estimation for Gradient-Boosted Regression Trees}. In \bibinfo{booktitle}{\emph{Advances in Neural Information Processing Systems}}, \bibfield{editor}{\bibinfo{person}{S.~Koyejo}, \bibinfo{person}{S.~Mohamed}, \bibinfo{person}{A.~Agarwal}, \bibinfo{person}{D.~Belgrave}, \bibinfo{person}{K.~Cho}, {and} \bibinfo{person}{A.~Oh}} (Eds.), Vol.~\bibinfo{volume}{35}. \bibinfo{publisher}{Curran Associates, Inc.}, \bibinfo{pages}{11145--11159}.
\newblock
\urldef\tempurl%
\url{https://proceedings.neurips.cc/paper_files/paper/2022/file/48088756ec0ce6ba362bddc7ebeb3915-Paper-Conference.pdf}
\showURL{%
\tempurl}


\bibitem[Bughin et~al\mbox{.}(2018)]%
        {bughin2018notes}
\bibfield{author}{\bibinfo{person}{Jacques Bughin}, \bibinfo{person}{Jeongmin Seong}, \bibinfo{person}{James Manyika}, \bibinfo{person}{Michael Chui}, {and} \bibinfo{person}{Raoul Joshi}.} \bibinfo{year}{2018}\natexlab{}.
\newblock \showarticletitle{Notes from the AI frontier: Modeling the impact of AI on the world economy}.
\newblock \bibinfo{journal}{\emph{McKinsey Global Institute}}  \bibinfo{volume}{4} (\bibinfo{year}{2018}).
\newblock


\bibitem[Chen and Guestrin(2016)]%
        {chen2016xgboost}
\bibfield{author}{\bibinfo{person}{Tianqi Chen} {and} \bibinfo{person}{Carlos Guestrin}.} \bibinfo{year}{2016}\natexlab{}.
\newblock \showarticletitle{Xgboost: A scalable tree boosting system}. In \bibinfo{booktitle}{\emph{Proceedings of the 22nd acm sigkdd international conference on knowledge discovery and data mining}}. \bibinfo{pages}{785--794}.
\newblock


\bibitem[Cohn et~al\mbox{.}(1996)]%
        {cohn1996active}
\bibfield{author}{\bibinfo{person}{David~A Cohn}, \bibinfo{person}{Zoubin Ghahramani}, {and} \bibinfo{person}{Michael~I Jordan}.} \bibinfo{year}{1996}\natexlab{}.
\newblock \showarticletitle{Active learning with statistical models}.
\newblock \bibinfo{journal}{\emph{Journal of artificial intelligence research}}  \bibinfo{volume}{4} (\bibinfo{year}{1996}), \bibinfo{pages}{129--145}.
\newblock


\bibitem[Duan et~al\mbox{.}(2020)]%
        {duan2020ngboost}
\bibfield{author}{\bibinfo{person}{Tony Duan}, \bibinfo{person}{Avati Anand}, \bibinfo{person}{Daisy~Yi Ding}, \bibinfo{person}{Khanh~K Thai}, \bibinfo{person}{Sanjay Basu}, \bibinfo{person}{Andrew Ng}, {and} \bibinfo{person}{Alejandro Schuler}.} \bibinfo{year}{2020}\natexlab{}.
\newblock \showarticletitle{Ngboost: Natural gradient boosting for probabilistic prediction}. In \bibinfo{booktitle}{\emph{International conference on machine learning}}. PMLR, \bibinfo{pages}{2690--2700}.
\newblock


\bibitem[Elreedy et~al\mbox{.}(2019)]%
        {elreedy2019novel}
\bibfield{author}{\bibinfo{person}{Dina Elreedy}, \bibinfo{person}{Amir F.~Atiya}, {and} \bibinfo{person}{Samir I.~Shaheen}.} \bibinfo{year}{2019}\natexlab{}.
\newblock \showarticletitle{A novel active learning regression framework for balancing the exploration-exploitation trade-off}.
\newblock \bibinfo{journal}{\emph{Entropy}} \bibinfo{volume}{21}, \bibinfo{number}{7} (\bibinfo{year}{2019}), \bibinfo{pages}{651}.
\newblock


\bibitem[Friedman(2002)]%
        {friedman2002stochastic}
\bibfield{author}{\bibinfo{person}{Jerome~H Friedman}.} \bibinfo{year}{2002}\natexlab{}.
\newblock \showarticletitle{Stochastic gradient boosting}.
\newblock \bibinfo{journal}{\emph{Computational statistics \& data analysis}} \bibinfo{volume}{38}, \bibinfo{number}{4} (\bibinfo{year}{2002}), \bibinfo{pages}{367--378}.
\newblock


\bibitem[Gal and Ghahramani(2015)]%
        {gal2015bayesian}
\bibfield{author}{\bibinfo{person}{Yarin Gal} {and} \bibinfo{person}{Zoubin Ghahramani}.} \bibinfo{year}{2015}\natexlab{}.
\newblock \showarticletitle{Bayesian convolutional neural networks with Bernoulli approximate variational inference}.
\newblock \bibinfo{journal}{\emph{arXiv preprint arXiv:1506.02158}} (\bibinfo{year}{2015}).
\newblock


\bibitem[Gal and Ghahramani(2016)]%
        {gal2016dropout}
\bibfield{author}{\bibinfo{person}{Yarin Gal} {and} \bibinfo{person}{Zoubin Ghahramani}.} \bibinfo{year}{2016}\natexlab{}.
\newblock \showarticletitle{Dropout as a bayesian approximation: Representing model uncertainty in deep learning}. In \bibinfo{booktitle}{\emph{international conference on machine learning}}. PMLR, \bibinfo{pages}{1050--1059}.
\newblock


\bibitem[Gal et~al\mbox{.}(2017)]%
        {gal2017deep}
\bibfield{author}{\bibinfo{person}{Yarin Gal}, \bibinfo{person}{Riashat Islam}, {and} \bibinfo{person}{Zoubin Ghahramani}.} \bibinfo{year}{2017}\natexlab{}.
\newblock \showarticletitle{Deep bayesian active learning with image data}. In \bibinfo{booktitle}{\emph{International conference on machine learning}}. PMLR, \bibinfo{pages}{1183--1192}.
\newblock


\bibitem[Grinsztajn et~al\mbox{.}(2022)]%
        {grinsztajn2022tree}
\bibfield{author}{\bibinfo{person}{L{\'e}o Grinsztajn}, \bibinfo{person}{Edouard Oyallon}, {and} \bibinfo{person}{Ga{\"e}l Varoquaux}.} \bibinfo{year}{2022}\natexlab{}.
\newblock \showarticletitle{Why do tree-based models still outperform deep learning on typical tabular data?}
\newblock \bibinfo{journal}{\emph{Advances in Neural Information Processing Systems}}  \bibinfo{volume}{35} (\bibinfo{year}{2022}), \bibinfo{pages}{507--520}.
\newblock


\bibitem[Holub et~al\mbox{.}(2008)]%
        {holub2008entropy}
\bibfield{author}{\bibinfo{person}{Alex Holub}, \bibinfo{person}{Pietro Perona}, {and} \bibinfo{person}{Michael~C Burl}.} \bibinfo{year}{2008}\natexlab{}.
\newblock \showarticletitle{Entropy-based active learning for object recognition}. In \bibinfo{booktitle}{\emph{2008 IEEE Computer Society Conference on Computer Vision and Pattern Recognition Workshops}}. IEEE, \bibinfo{pages}{1--8}.
\newblock


\bibitem[Holzm{\"u}ller et~al\mbox{.}(2023)]%
        {holzmuller2023framework}
\bibfield{author}{\bibinfo{person}{David Holzm{\"u}ller}, \bibinfo{person}{Viktor Zaverkin}, \bibinfo{person}{Johannes K{\"a}stner}, {and} \bibinfo{person}{Ingo Steinwart}.} \bibinfo{year}{2023}\natexlab{}.
\newblock \showarticletitle{A framework and benchmark for deep batch active learning for regression}.
\newblock \bibinfo{journal}{\emph{Journal of Machine Learning Research}} \bibinfo{volume}{24}, \bibinfo{number}{164} (\bibinfo{year}{2023}), \bibinfo{pages}{1--81}.
\newblock


\bibitem[Houlsby et~al\mbox{.}(2011)]%
        {houlsby2011bayesian}
\bibfield{author}{\bibinfo{person}{Neil Houlsby}, \bibinfo{person}{Ferenc Husz{\'a}r}, \bibinfo{person}{Zoubin Ghahramani}, {and} \bibinfo{person}{M{\'a}t{\'e} Lengyel}.} \bibinfo{year}{2011}\natexlab{}.
\newblock \showarticletitle{Bayesian active learning for classification and preference learning}.
\newblock \bibinfo{journal}{\emph{arXiv preprint arXiv:1112.5745}} (\bibinfo{year}{2011}).
\newblock


\bibitem[H{\"u}llermeier and Waegeman(2021)]%
        {hullermeier2021aleatoric}
\bibfield{author}{\bibinfo{person}{Eyke H{\"u}llermeier} {and} \bibinfo{person}{Willem Waegeman}.} \bibinfo{year}{2021}\natexlab{}.
\newblock \showarticletitle{Aleatoric and epistemic uncertainty in machine learning: An introduction to concepts and methods}.
\newblock \bibinfo{journal}{\emph{Machine Learning}}  \bibinfo{volume}{110} (\bibinfo{year}{2021}), \bibinfo{pages}{457--506}.
\newblock


\bibitem[Kagy et~al\mbox{.}(2019)]%
        {kagy2019practical}
\bibfield{author}{\bibinfo{person}{Jean-Fran{\c{c}}ois Kagy}, \bibinfo{person}{Tolga Kayadelen}, \bibinfo{person}{Ji Ma}, \bibinfo{person}{Afshin Rostamizadeh}, {and} \bibinfo{person}{Jana Strnadova}.} \bibinfo{year}{2019}\natexlab{}.
\newblock \showarticletitle{The practical challenges of active learning: lessons learned from live experimentation}.
\newblock \bibinfo{journal}{\emph{arXiv preprint arXiv:1907.00038}} (\bibinfo{year}{2019}).
\newblock


\bibitem[Katzir et~al\mbox{.}(2020)]%
        {katzir2020net}
\bibfield{author}{\bibinfo{person}{Liran Katzir}, \bibinfo{person}{Gal Elidan}, {and} \bibinfo{person}{Ran El-Yaniv}.} \bibinfo{year}{2020}\natexlab{}.
\newblock \showarticletitle{Net-dnf: Effective deep modeling of tabular data}. In \bibinfo{booktitle}{\emph{International conference on learning representations}}.
\newblock


\bibitem[Ke et~al\mbox{.}(2017)]%
        {ke2017lightgbm}
\bibfield{author}{\bibinfo{person}{Guolin Ke}, \bibinfo{person}{Qi Meng}, \bibinfo{person}{Thomas Finley}, \bibinfo{person}{Taifeng Wang}, \bibinfo{person}{Wei Chen}, \bibinfo{person}{Weidong Ma}, \bibinfo{person}{Qiwei Ye}, {and} \bibinfo{person}{Tie-Yan Liu}.} \bibinfo{year}{2017}\natexlab{}.
\newblock \showarticletitle{Lightgbm: A highly efficient gradient boosting decision tree}.
\newblock \bibinfo{journal}{\emph{Advances in neural information processing systems}}  \bibinfo{volume}{30} (\bibinfo{year}{2017}), \bibinfo{pages}{3146--3154}.
\newblock


\bibitem[Kendall and Gal(2017)]%
        {kendall2017uncertainties}
\bibfield{author}{\bibinfo{person}{Alex Kendall} {and} \bibinfo{person}{Yarin Gal}.} \bibinfo{year}{2017}\natexlab{}.
\newblock \showarticletitle{What uncertainties do we need in bayesian deep learning for computer vision?}
\newblock \bibinfo{journal}{\emph{Advances in neural information processing systems}}  \bibinfo{volume}{30} (\bibinfo{year}{2017}).
\newblock


\bibitem[Kirsch et~al\mbox{.}(2019)]%
        {kirsch2019batchbald}
\bibfield{author}{\bibinfo{person}{Andreas Kirsch}, \bibinfo{person}{Joost Van~Amersfoort}, {and} \bibinfo{person}{Yarin Gal}.} \bibinfo{year}{2019}\natexlab{}.
\newblock \showarticletitle{Batchbald: Efficient and diverse batch acquisition for deep bayesian active learning}.
\newblock \bibinfo{journal}{\emph{Advances in neural information processing systems}}  \bibinfo{volume}{32} (\bibinfo{year}{2019}).
\newblock


\bibitem[Lewis(1995)]%
        {lewis1995sequential}
\bibfield{author}{\bibinfo{person}{David~D Lewis}.} \bibinfo{year}{1995}\natexlab{}.
\newblock \showarticletitle{A sequential algorithm for training text classifiers: Corrigendum and additional data}. In \bibinfo{booktitle}{\emph{Acm Sigir Forum}}, Vol.~\bibinfo{volume}{29}. ACM New York, NY, USA, \bibinfo{pages}{13--19}.
\newblock


\bibitem[Lowell et~al\mbox{.}(2019)]%
        {lowell2019practical}
\bibfield{author}{\bibinfo{person}{David Lowell}, \bibinfo{person}{Zachary~C Lipton}, {and} \bibinfo{person}{Byron~C Wallace}.} \bibinfo{year}{2019}\natexlab{}.
\newblock \showarticletitle{Practical Obstacles to Deploying Active Learning}. In \bibinfo{booktitle}{\emph{Proceedings of the 2019 Conference on Empirical Methods in Natural Language Processing and the 9th International Joint Conference on Natural Language Processing (EMNLP-IJCNLP)}}. \bibinfo{pages}{21--30}.
\newblock


\bibitem[Malinin et~al\mbox{.}(2021)]%
        {malinin2021uncertainty}
\bibfield{author}{\bibinfo{person}{Andrey Malinin}, \bibinfo{person}{Liudmila Prokhorenkova}, {and} \bibinfo{person}{Aleksei Ustimenko}.} \bibinfo{year}{2021}\natexlab{}.
\newblock \showarticletitle{Uncertainty in Gradient Boosting via Ensembles}. In \bibinfo{booktitle}{\emph{International Conference on Learning Representations}}.
\newblock
\urldef\tempurl%
\url{https://openreview.net/forum?id=1Jv6b0Zq3qi}
\showURL{%
\tempurl}


\bibitem[Munro and Monarch(2021)]%
        {munro2021human}
\bibfield{author}{\bibinfo{person}{Robert Munro} {and} \bibinfo{person}{Robert Monarch}.} \bibinfo{year}{2021}\natexlab{}.
\newblock \bibinfo{booktitle}{\emph{Human-in-the-Loop Machine Learning: Active learning and annotation for human-centered AI}}.
\newblock \bibinfo{publisher}{Simon and Schuster}.
\newblock


\bibitem[Nguyen et~al\mbox{.}(2019)]%
        {nguyen2019epistemic}
\bibfield{author}{\bibinfo{person}{Vu-Linh Nguyen}, \bibinfo{person}{S{\'e}bastien Destercke}, {and} \bibinfo{person}{Eyke H{\"u}llermeier}.} \bibinfo{year}{2019}\natexlab{}.
\newblock \showarticletitle{Epistemic uncertainty sampling}. In \bibinfo{booktitle}{\emph{Discovery Science: 22nd International Conference, DS 2019, Split, Croatia, October 28--30, 2019, Proceedings 22}}. Springer, \bibinfo{pages}{72--86}.
\newblock


\bibitem[Nguyen et~al\mbox{.}(2022)]%
        {nguyen2022measure}
\bibfield{author}{\bibinfo{person}{Vu-Linh Nguyen}, \bibinfo{person}{Mohammad~Hossein Shaker}, {and} \bibinfo{person}{Eyke H{\"u}llermeier}.} \bibinfo{year}{2022}\natexlab{}.
\newblock \showarticletitle{How to measure uncertainty in uncertainty sampling for active learning}.
\newblock \bibinfo{journal}{\emph{Machine Learning}} \bibinfo{volume}{111}, \bibinfo{number}{1} (\bibinfo{year}{2022}), \bibinfo{pages}{89--122}.
\newblock


\bibitem[Pace and Barry({[n.\,d.]})]%
        {california_housing}
\bibfield{author}{\bibinfo{person}{R.~Kelley Pace} {and} \bibinfo{person}{Ronald Barry}.} \bibinfo{year}{[n.\,d.]}\natexlab{}.
\newblock  (\bibinfo{year}{[n.\,d.]}).
\newblock


\bibitem[Prokhorenkova et~al\mbox{.}(2018)]%
        {prokhorenkova2018catboost}
\bibfield{author}{\bibinfo{person}{Liudmila Prokhorenkova}, \bibinfo{person}{Gleb Gusev}, \bibinfo{person}{Aleksandr Vorobev}, \bibinfo{person}{Anna~Veronika Dorogush}, {and} \bibinfo{person}{Andrey Gulin}.} \bibinfo{year}{2018}\natexlab{}.
\newblock \showarticletitle{CatBoost: unbiased boosting with categorical features}.
\newblock \bibinfo{journal}{\emph{Advances in neural information processing systems}}  \bibinfo{volume}{31} (\bibinfo{year}{2018}).
\newblock


\bibitem[Quinlan(1987)]%
        {misc_thyroid_disease_102}
\bibfield{author}{\bibinfo{person}{Ross Quinlan}.} \bibinfo{year}{1987}\natexlab{}.
\newblock \bibinfo{title}{{Thyroid Disease}}.
\newblock \bibinfo{howpublished}{UCI Machine Learning Repository}.
\newblock
\newblock
\shownote{{DOI}: https://doi.org/10.24432/C5D010}.


\bibitem[Redmond(2009)]%
        {communities_and_crime}
\bibfield{author}{\bibinfo{person}{Michael Redmond}.} \bibinfo{year}{2009}\natexlab{}.
\newblock \bibinfo{title}{{Communities and Crime}}.
\newblock \bibinfo{howpublished}{UCI Machine Learning Repository}.
\newblock
\newblock
\shownote{{DOI}: https://doi.org/10.24432/C53W3X}.


\bibitem[Ren et~al\mbox{.}(2021)]%
        {ren2021survey}
\bibfield{author}{\bibinfo{person}{Pengzhen Ren}, \bibinfo{person}{Yun Xiao}, \bibinfo{person}{Xiaojun Chang}, \bibinfo{person}{Po-Yao Huang}, \bibinfo{person}{Zhihui Li}, \bibinfo{person}{Brij~B Gupta}, \bibinfo{person}{Xiaojiang Chen}, {and} \bibinfo{person}{Xin Wang}.} \bibinfo{year}{2021}\natexlab{}.
\newblock \showarticletitle{A survey of deep active learning}.
\newblock \bibinfo{journal}{\emph{ACM computing surveys (CSUR)}} \bibinfo{volume}{54}, \bibinfo{number}{9} (\bibinfo{year}{2021}), \bibinfo{pages}{1--40}.
\newblock


\bibitem[Scheffer et~al\mbox{.}(2001)]%
        {scheffer2001active}
\bibfield{author}{\bibinfo{person}{Tobias Scheffer}, \bibinfo{person}{Christian Decomain}, {and} \bibinfo{person}{Stefan Wrobel}.} \bibinfo{year}{2001}\natexlab{}.
\newblock \showarticletitle{Active hidden markov models for information extraction}. In \bibinfo{booktitle}{\emph{International symposium on intelligent data analysis}}. Springer, \bibinfo{pages}{309--318}.
\newblock


\bibitem[Senge et~al\mbox{.}(2014)]%
        {senge2014reliable}
\bibfield{author}{\bibinfo{person}{Robin Senge}, \bibinfo{person}{Stefan B{\"o}sner}, \bibinfo{person}{Krzysztof Dembczy{\'n}ski}, \bibinfo{person}{J{\"o}rg Haasenritter}, \bibinfo{person}{Oliver Hirsch}, \bibinfo{person}{Norbert Donner-Banzhoff}, {and} \bibinfo{person}{Eyke H{\"u}llermeier}.} \bibinfo{year}{2014}\natexlab{}.
\newblock \showarticletitle{Reliable classification: Learning classifiers that distinguish aleatoric and epistemic uncertainty}.
\newblock \bibinfo{journal}{\emph{Information Sciences}}  \bibinfo{volume}{255} (\bibinfo{year}{2014}), \bibinfo{pages}{16--29}.
\newblock


\bibitem[Settles(2009)]%
        {settles2009active}
\bibfield{author}{\bibinfo{person}{Burr Settles}.} \bibinfo{year}{2009}\natexlab{}.
\newblock \showarticletitle{Active learning literature survey}.
\newblock  (\bibinfo{year}{2009}).
\newblock


\bibitem[Settles and Craven(2008)]%
        {settles2008analysis}
\bibfield{author}{\bibinfo{person}{Burr Settles} {and} \bibinfo{person}{Mark Craven}.} \bibinfo{year}{2008}\natexlab{}.
\newblock \showarticletitle{An analysis of active learning strategies for sequence labeling tasks}. In \bibinfo{booktitle}{\emph{proceedings of the 2008 conference on empirical methods in natural language processing}}. \bibinfo{pages}{1070--1079}.
\newblock


\bibitem[Shaker and Hüllermeier(2020)]%
        {Shaker_2020}
\bibfield{author}{\bibinfo{person}{Mohammad~Hossein Shaker} {and} \bibinfo{person}{Eyke Hüllermeier}.} \bibinfo{year}{2020}\natexlab{}.
\newblock \showarticletitle{Aleatoric and Epistemic Uncertainty with Random Forests}.
\newblock In \bibinfo{booktitle}{\emph{Lecture Notes in Computer Science}}. \bibinfo{publisher}{Springer International Publishing}, \bibinfo{pages}{444--456}.
\newblock
\urldef\tempurl%
\url{https://doi.org/10.1007/978-3-030-44584-3_35}
\showDOI{\tempurl}


\bibitem[Shannon(1948)]%
        {shannon1948mathematical}
\bibfield{author}{\bibinfo{person}{Claude~Elwood Shannon}.} \bibinfo{year}{1948}\natexlab{}.
\newblock \showarticletitle{A mathematical theory of communication}.
\newblock \bibinfo{journal}{\emph{The Bell system technical journal}} \bibinfo{volume}{27}, \bibinfo{number}{3} (\bibinfo{year}{1948}), \bibinfo{pages}{379--423}.
\newblock


\bibitem[Shwartz-Ziv and Armon(2022)]%
        {shwartz2022tabular}
\bibfield{author}{\bibinfo{person}{Ravid Shwartz-Ziv} {and} \bibinfo{person}{Amitai Armon}.} \bibinfo{year}{2022}\natexlab{}.
\newblock \showarticletitle{Tabular data: Deep learning is not all you need}.
\newblock \bibinfo{journal}{\emph{Information Fusion}}  \bibinfo{volume}{81} (\bibinfo{year}{2022}), \bibinfo{pages}{84--90}.
\newblock


\bibitem[Tfekci and Kaya(2014)]%
        {energy}
\bibfield{author}{\bibinfo{person}{Pnar Tfekci} {and} \bibinfo{person}{Heysem Kaya}.} \bibinfo{year}{2014}\natexlab{}.
\newblock \bibinfo{title}{{Combined Cycle Power Plant}}.
\newblock \bibinfo{howpublished}{UCI Machine Learning Repository}.
\newblock
\newblock
\shownote{{DOI}: https://doi.org/10.24432/C5002N}.


\bibitem[Tsymbalov et~al\mbox{.}(2018)]%
        {tsymbalov2018dropout}
\bibfield{author}{\bibinfo{person}{Evgenii Tsymbalov}, \bibinfo{person}{Maxim Panov}, {and} \bibinfo{person}{Alexander Shapeev}.} \bibinfo{year}{2018}\natexlab{}.
\newblock \showarticletitle{Dropout-based active learning for regression}. In \bibinfo{booktitle}{\emph{Analysis of Images, Social Networks and Texts: 7th International Conference, AIST 2018, Moscow, Russia, July 5--7, 2018, Revised Selected Papers 7}}. Springer, \bibinfo{pages}{247--258}.
\newblock


\bibitem[Ustimenko and Prokhorenkova(2021)]%
        {ustimenko2021sglb}
\bibfield{author}{\bibinfo{person}{Aleksei Ustimenko} {and} \bibinfo{person}{Liudmila Prokhorenkova}.} \bibinfo{year}{2021}\natexlab{}.
\newblock \showarticletitle{SGLB: Stochastic gradient langevin boosting}. In \bibinfo{booktitle}{\emph{International Conference on Machine Learning}}. PMLR, \bibinfo{pages}{10487--10496}.
\newblock


\bibitem[Wang et~al\mbox{.}(2016)]%
        {wang2016cost}
\bibfield{author}{\bibinfo{person}{Keze Wang}, \bibinfo{person}{Dongyu Zhang}, \bibinfo{person}{Ya Li}, \bibinfo{person}{Ruimao Zhang}, {and} \bibinfo{person}{Liang Lin}.} \bibinfo{year}{2016}\natexlab{}.
\newblock \showarticletitle{Cost-effective active learning for deep image classification}.
\newblock \bibinfo{journal}{\emph{IEEE Transactions on Circuits and Systems for Video Technology}} \bibinfo{volume}{27}, \bibinfo{number}{12} (\bibinfo{year}{2016}), \bibinfo{pages}{2591--2600}.
\newblock


\bibitem[Welling and Teh(2011)]%
        {welling2011bayesian}
\bibfield{author}{\bibinfo{person}{Max Welling} {and} \bibinfo{person}{Yee~W Teh}.} \bibinfo{year}{2011}\natexlab{}.
\newblock \showarticletitle{Bayesian learning via stochastic gradient Langevin dynamics}. In \bibinfo{booktitle}{\emph{Proceedings of the 28th international conference on machine learning (ICML-11)}}. \bibinfo{pages}{681--688}.
\newblock


\bibitem[Wu et~al\mbox{.}(2019)]%
        {wu2019active}
\bibfield{author}{\bibinfo{person}{Dongrui Wu}, \bibinfo{person}{Chin-Teng Lin}, {and} \bibinfo{person}{Jian Huang}.} \bibinfo{year}{2019}\natexlab{}.
\newblock \showarticletitle{Active learning for regression using greedy sampling}.
\newblock \bibinfo{journal}{\emph{Information Sciences}}  \bibinfo{volume}{474} (\bibinfo{year}{2019}), \bibinfo{pages}{90--105}.
\newblock


\bibitem[Zhou et~al\mbox{.}(2022)]%
        {zhou2022survey}
\bibfield{author}{\bibinfo{person}{Xinlei Zhou}, \bibinfo{person}{Han Liu}, \bibinfo{person}{Farhad Pourpanah}, \bibinfo{person}{Tieyong Zeng}, {and} \bibinfo{person}{Xizhao Wang}.} \bibinfo{year}{2022}\natexlab{}.
\newblock \showarticletitle{A survey on epistemic (model) uncertainty in supervised learning: Recent advances and applications}.
\newblock \bibinfo{journal}{\emph{Neurocomputing}}  \bibinfo{volume}{489} (\bibinfo{year}{2022}), \bibinfo{pages}{449--465}.
\newblock


\end{thebibliography}

% \appendix





\end{document}